\documentclass{llncs} 

\usepackage{float}
\usepackage{graphicx}
\usepackage{epsfig}
\usepackage{multirow}
\usepackage{setspace}
\usepackage{amsmath,amssymb}
\usepackage{color}
\usepackage{mathptmx}

\begin{document}

\title{Active Canny: Edge Detection and Recovery with Open Active Contour Models}
\titlerunning{Edge Detection and Recovery with Open Active Contour Models}
\author{Muhammet Ba\c{s}tan \and S. Saqib Bukhari \and Thomas M. Breuel
\institute{Technical University of Kaiserslautern,\\Kaiserslautern, Germany\\ \email{mubastan@gmail.com, saqib.bukhari@dfki.de, tmb@cs.uni-kl.de}}}

\maketitle

\begin{abstract}

We introduce an edge detection and recovery framework based on open active contour models (snakelets). This is motivated by the noisy or broken edges output by standard edge detection algorithms, like Canny. The idea is to utilize the local continuity and smoothness cues provided by strong edges and grow them to recover the missing edges. This way, the strong edges are used to recover weak or missing edges by considering the local edge structures, instead of blindly linking them if gradient magnitudes are above some threshold. We initialize short snakelets on the gradient magnitudes or binary edges automatically and then deform and grow them under the influence of gradient vector flow. The output snakelets are able to recover most of the breaks or weak edges, and they provide a smooth edge representation of the image; they can also be used for higher level analysis, like contour segmentation.

\end{abstract}

\section{Introduction}

Edge detection is a fundamental tool in computer vision, especially for feature detection and extraction.
In an image, edges are significant local changes in the intensity values and 
they indicate the presence of a boundary between adjacent regions.
It is well known that edge detection in non-ideal images usually results in noisy edges, disconnected (broken) edges or both due to various reasons. The noise problem can be alleviated by using higher thresholds, but this in turn may deteriorate the problem of broken edges, as shown in Figure~\ref{fig:canny-edges}.
Short breaks in the edges can be recovered with simple post-processing (e.g., morphology), while large breaks need special treatment.

\begin{figure}[h!t]
    
    \centerline{Input \hspace{30mm} Output edges}
    \centerline{\epsfig{figure=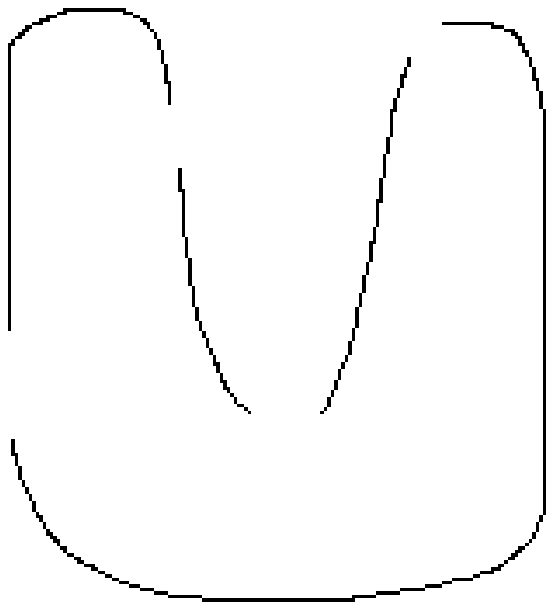,width=0.35\textwidth} 
		\epsfig{figure=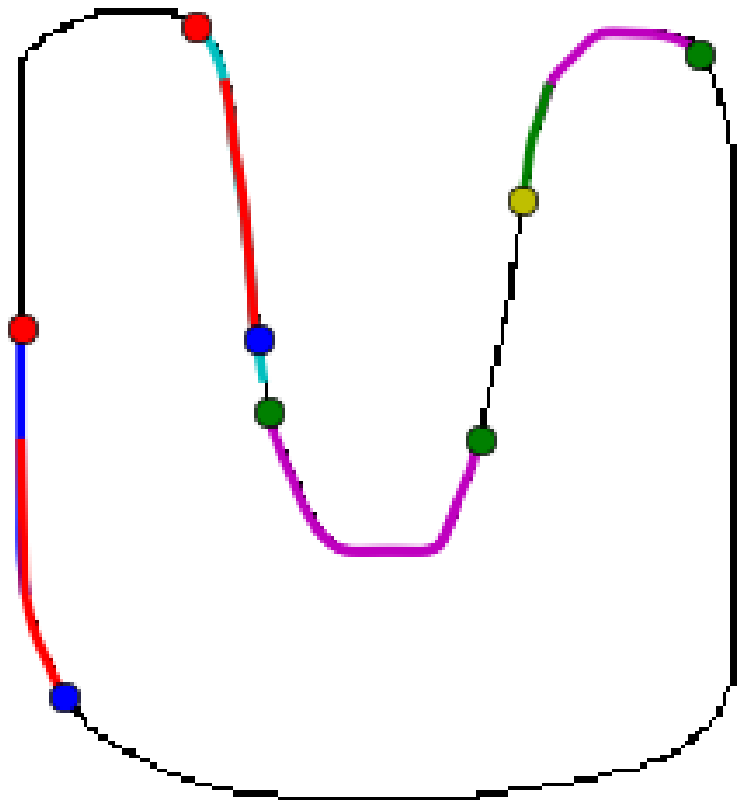,width=0.35\textwidth}}

    \centerline{\epsfig{figure=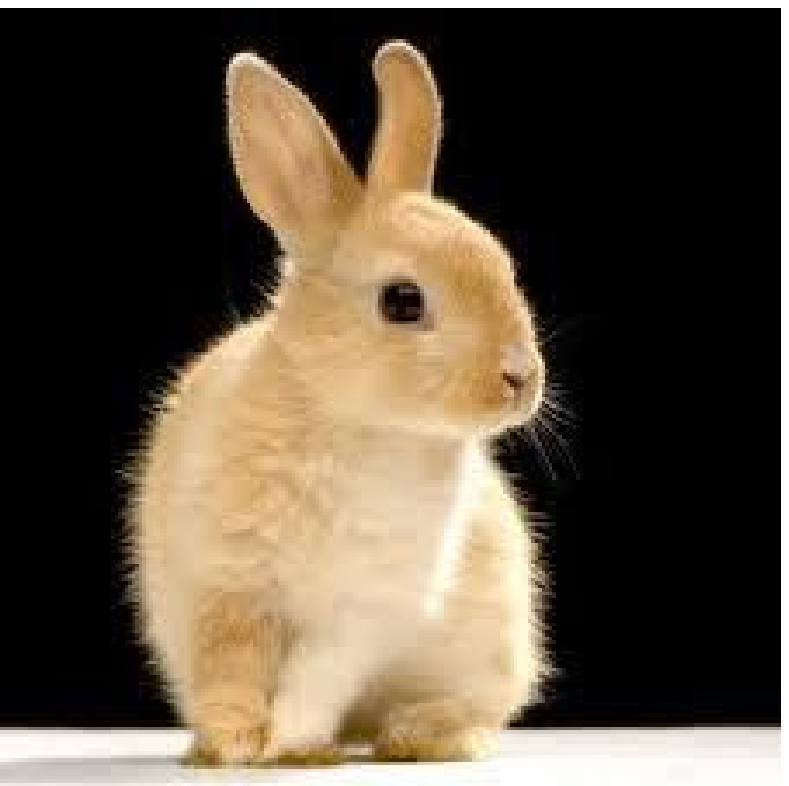,width=0.35\textwidth} 
		\epsfig{figure=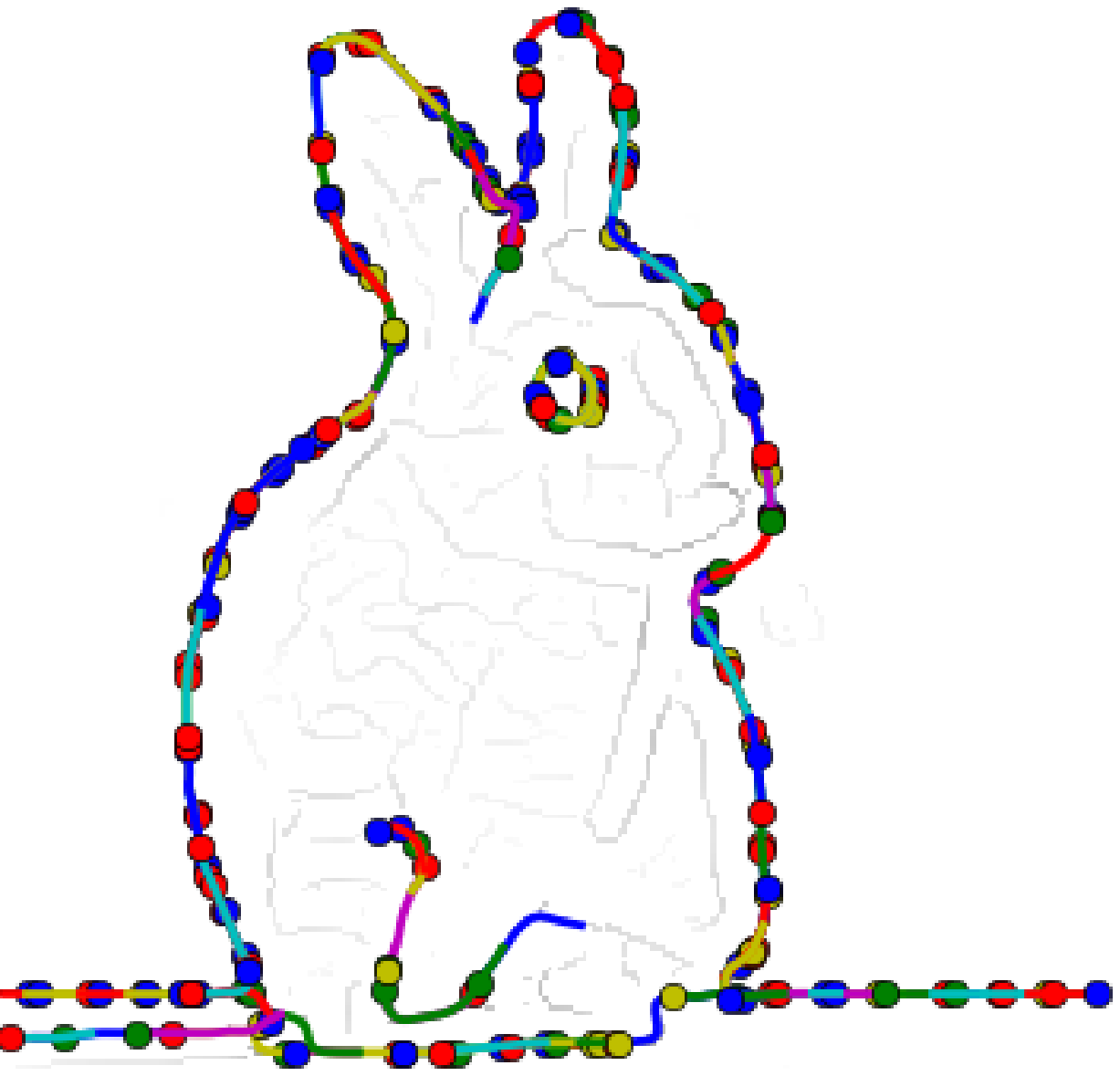,width=0.35\textwidth}}
		
    \caption{The snakelets can be used for edge recovery (first row) on binary edge images or edge detection (second row) on grayscale or color images.}    
    \label{fig:intro}
\end{figure}

In this paper, we address this problem and introduce an open active contour models (snakelets) for edge detection and recovery. In edge recovery, the input is a binary edge image possibly with breaks (missing edge pixels) and the goal is to recover the breaks (Figure~\ref{fig:intro}, first row). In edge detection, the input is a grayscale or color image and the goal is to detect all edges without noise (Figure~\ref{fig:intro}, second row). The idea is to utilize the local edge shape and gradient magnitudes with the help of snakelets and connect edge/contour pixels, instead of blindly connecting them based only on thresholds, as done, for instance, in Canny edge detection~\cite{canny-pami-86}.

To summarize our framework, we initialize short snakelets on the gradient magnitude or binary edge images, and then deform and grow them along the edges under the influence of gradient vector flow (GVF)~\cite{xu98}. For example, in Figure~\ref{fig:intro}, first row, we initialized snakelets on a binary edge image with some breaks and obtained the output on the right. In the second row, we did edge detection on an RGB color image, by initializing snakelets on gradient magnitudes.

Open active contour models have been successfully employed in different domains, especially in medical image analysis, for extracting filament or tree-like tubular structures~\cite{Hongsheng-Li-2009,Hongsheng-Li-2009a,Ting-Xu-2011,Yu-Wang-2011,Yu-Wan-2011b}; and in document image analysis for extracting curved text-lines, which also look like filaments, from document images~\cite{bukharidas08,bukhari-ijdar-2011}.

Li et al.~\cite{Hongsheng-Li-2009} introduced stretching open active contours (SOAC) model for tracking and elongation measurement of actin filaments, which is one of the well-known problems in medical image processing. The SOAC model is an advancement of the author's previous work~\cite{Hongsheng-Li-2009a}, which mainly had snake over-growing and/or under-growing problems near filament tips.
The SOAC model overcomes this problem by adding non-intensity-adaptive stretching term, which also naturally adds continuous body constraints over a filament body.

An open-source software tool for open-curve active contour model is presented by Smith et al.~\cite{smith-cs-2010}.
This tool is based on SOAC model~\cite{Hongsheng-Li-2009}, and it can be used interactively for segmenting and tracking filamentous structures.

Xu et al.~\cite{Ting-Xu-2011} introduced a method for extracting an intersecting and/or overlapping actin filaments network by automatic initialization of multiple SOAC models. For this purpose, they used normalized cuts for dissecting and then re-grouping SOAC segments.

Wang et al.~\cite{Yu-Wang-2011} introduced a method for neuron tracing, based on a combination of a 3-D open-curve active contour tracing model and a set of novel preprocessing and postprocessing rules.
In contrast to other open-curve snake based neuron tracing methods, their method is a generic framework and can be applied to a variety of 3-D neuron images.

A 4-D open-curve active contour model is introduced by Wang et al.~\cite{Yu-Wan-2011b}, based on 3-D centerline tracing and local radius estimation at the same time. This model is presented for extraction of tree-like tubular structures from stacks of 3-D images.

An active contour based `baby-snakes' curled text-line segmentation algorithm is introduced by Bukhari et al.~\cite{bukharidas08}. 
In this method, an input image is first smeared by using morphological operations.
Open-curve slope-aligned snakes are initialized over the smeared connected components.
After deformation, each group of joined snakes is considered as a segmented text-line.

Bukhari et al.~\cite{bukhari-ijdar-2011} also introduced another active contour based snakelets model for curled text-line finding.
In this method, a pair of straight open-curve snakes is initialized over each connected component's top and bottom points.
The top and bottom snakes are deformed using the external energies of neighboring top and bottom points, respectively.
Each overlapping pairs of snakes is returned as a segmented text-line.

Our snakelet model is similar to those mentioned above, but for solving a more fundamental problem, edge detection and recovery.
There are also some work on ``contour completion''~\cite{contour-iccv05,contour-ijcv08,contour-cvpr12}, in which the goal is to recover missing contours and/or obtain closed contours using the available edge/contour fragments. These train a high-level model, e.g., with conditional random fields (CRFs), on a set of labeled images. Compared to these, our work can be considered a low level approach, without any high-level or global modeling and does not need training. We believe that the output of our system can be used in these high level contour completion frameworks to obtain better results.

\section{Canny Edge Detection}

In this section we review the Canny edge detection algorithm~\cite{canny-pami-86,contour-review-ivc11}, which is a widely used edge detection algorithm.
Canny's approach is based on three objectives: (1) low error rate, (2) good localization of the edge points, (3) thin edges. It is a four stage algorithm: (1) smooth the image with a Gaussian filter for noise reduction, (2) compute the gradient magnitudes and angles, (3) apply nonmaxima suppression to the gradient magnitudes based on the angles, (4) use hysteresis thresholding and link the edges. The algorithm is originally designed for grayscale images, but later adapted to color multi channel images (e.g.,~\cite{weijer-tip-2006}).

\begin{figure}[h!t]    
    \centerline{\epsfig{figure=figures/bunny.eps,width=0.22\textwidth} 
                \epsfig{figure=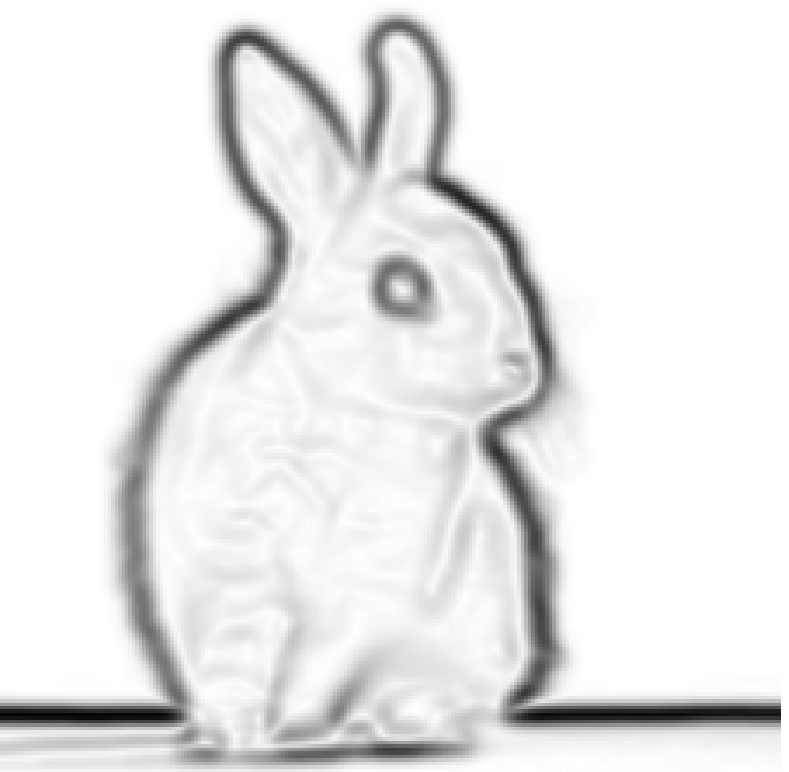,width=0.22\textwidth}      
		\epsfig{figure=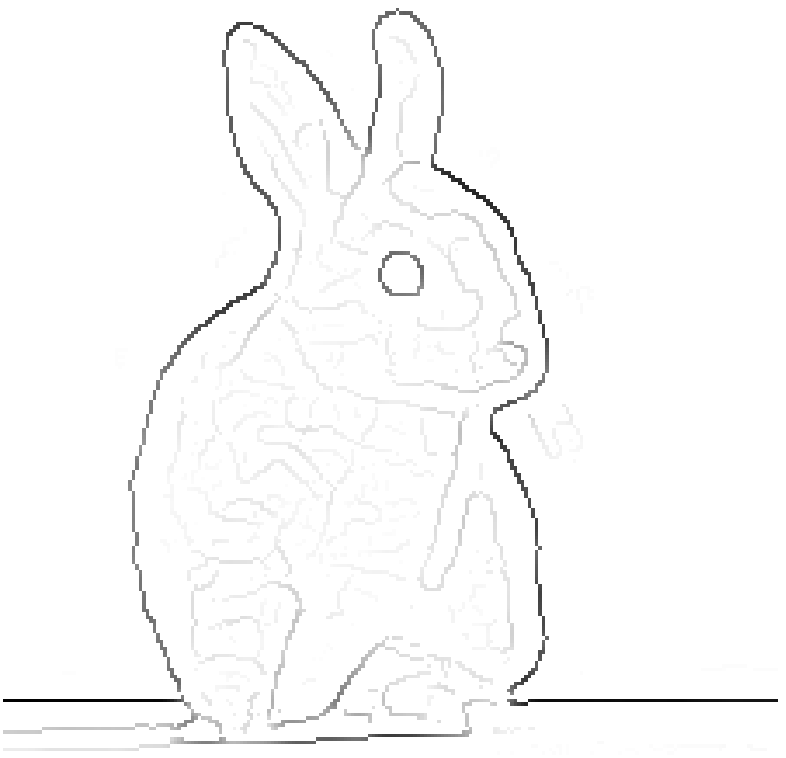,width=0.22\textwidth}}
    \centerline{\epsfig{figure=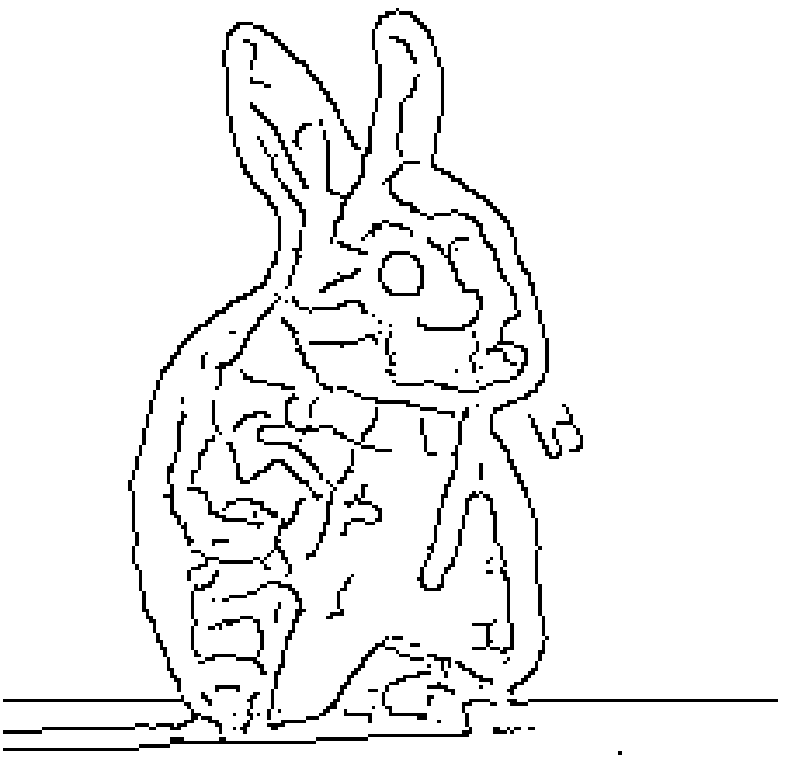,width=0.22\textwidth} 
                \epsfig{figure=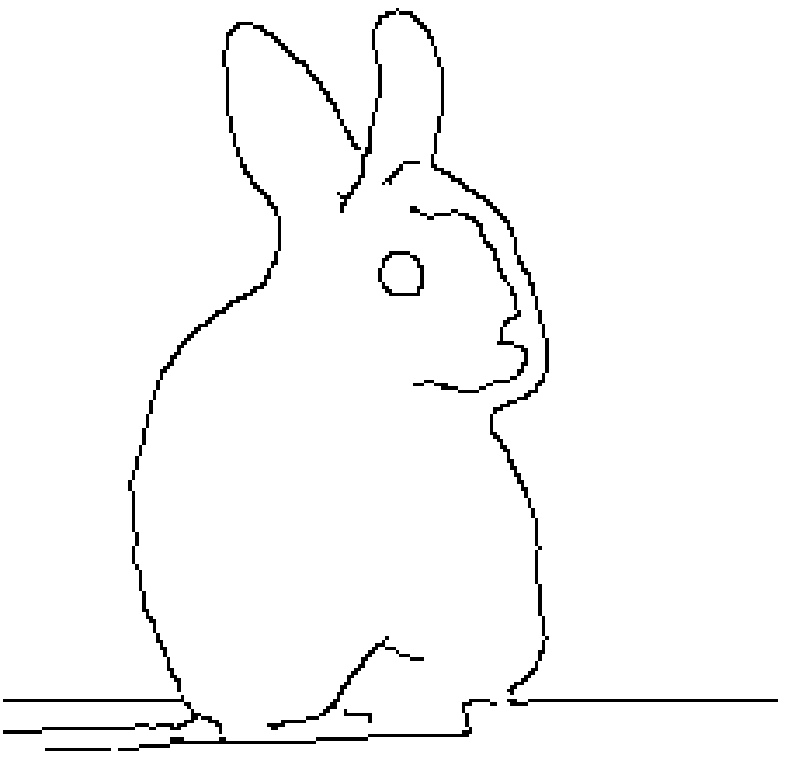,width=0.22\textwidth}      
		\epsfig{figure=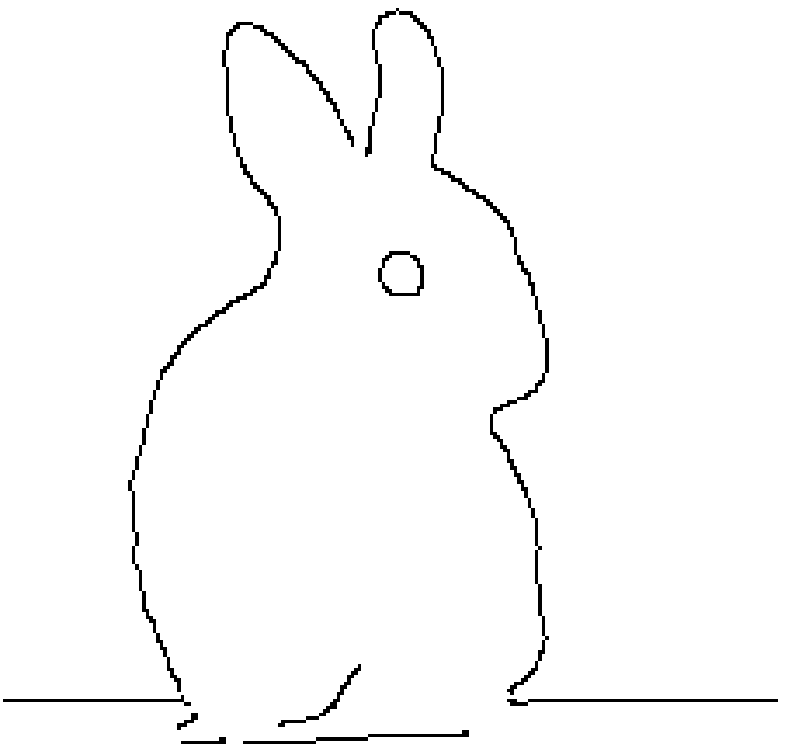,width=0.22\textwidth}}    
    
    \caption{Canny edge detection on an RGB color image. Top row: color image, color gradient magnitude (\cite{weijer-tip-2006}), gradient magnitude after nonmaximum suppression. Bottom: Canny edge detection results using low, medium and high thresholds. Even low thresholds result in breaks in the edges, since the edge magnitudes are very low or zero due to illumination/noise.}
    \label{fig:canny-edges}
\end{figure}

The performance of the Canny algorithm is very much dependent on the high and low thresholds (TH, TL) used in the hysteresis thresholding stage. When the thresholds are low, we get noisy edges; when they are high, we miss true edges and contours become disconnected (broken edges).
We may still get broken edges in some cases (e.g., noise, ramp edges) even if we set the thresholds low.
Figure~\ref{fig:canny-edges} demonstrates this well-known problem.

The thresholds can be estimated from the gradient magnitudes based on the mean, median, p-fractile or similar statistics, however, these estimated thresholds do not generalize well.
Our motivation is to use high thresholds and exploit the cues in the local edge structures to extend the strong edges.
The edge linking step in Canny is blind and does not make use of the continuity and smoothness cues available in the already available strong edges to link them with the weak edges.

\section{The Snakelet Model}
Our snakelets model is the same as the standard snakelet models used in the literature~\cite{smith-cs-2010,Yu-Wang-2011}, with minor differences to adapt to our framework.
Like a traditional snake model~\cite{kass88}, our snakelets model minimizes internal and external energy; the internal energy controls the shape of the snakelets, the external energy attracts the snakelets towards edges and simultaneously grows them along the edges.

\begin{figure}[h!t]
    \centerline{\epsfig{figure=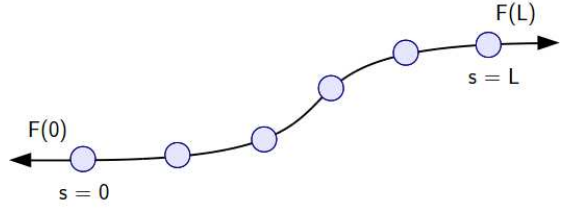,width=0.60\textwidth}}    
    \caption{Our snakelet model consists of uniformly sampled points along the active contour and growing forces at one or both ends (unidirectional/bidirectional snakelet).}
    \label{fig:snakelet}
\end{figure}

A snakelet in 2-D is represented by a set of points $S(s) = [x(s), y(s)]$ (Figure~\ref{fig:snakelet}), where $s\in[0, 1]$, that
moves through the spatial domain of an image to minimize the energy function ($E$) as shown in Equation~\ref{snk:eq1}.

\begin{equation}
\label{snk:eq1}
E=\int_0^1{[E_{int}\{S(s)\}+E_{ext}\{S(s)\}]ds}
\end{equation}

The snakelet slithers towards a targeted object under the 
influence of internal energy ($E_{int}$) and external energy ($E_{ext}$), 
where the internal energy is estimated from the snakelet points 
and the external energy is computed from the image contents. 

\subsection{The Internal Energy}
The internal energy ($E_{int}$) is defined similar to a traditional closed curve snake~\cite{kass88} which is shown in Equation~\ref{snk:eq2}, 
where it is decomposed into two factors: (i) $S'(s)$ (first order derivative of $S(s)$) represents tension,
(ii) $S''(s)$ (second order derivative of $S(s)$) represents the rigidity of the snakelet.
The weight parameters $\alpha$ and $\beta$ are used for controlling snakelet's tension and rigidity, respectively. 
The tension tries to shrink the snakelet, and since our snakelets should grow we set the value of $\alpha(s)$ to small values.
Due to open curve property, we set the value of $\beta(s)$ to 0 for $s=0$ and $s=1$.

\begin{equation}
\label{snk:eq2}
E_{int}=\int_0^1{[\alpha \{S'(s)\}+\beta
\{S''(s)\}]ds}
\end{equation}

\subsection{The External Energy}
The external energy is composed of two terms: the edge term and the growing term as shown in 
Equation~\ref{snk:eq3}.

\begin{equation}
\label{snk:eq3}
E_{ext}=\int_0^1{[ E_{edge}\{S(s)\} + \gamma E_{grow}\{S(s)\} ] ds}
\end{equation}

\begin{figure}[h!t]
    \centerline{Input edge image}
    \centerline{\epsfig{figure=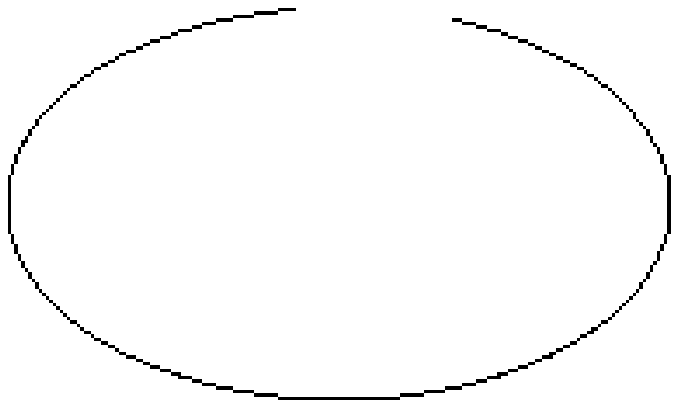,width=0.33\textwidth}}
    
    \centerline{ \hspace{15mm} 3 iterations \hspace{30mm}  10 iterations}
    \centerline{GVF x \hspace{3mm} \epsfig{figure=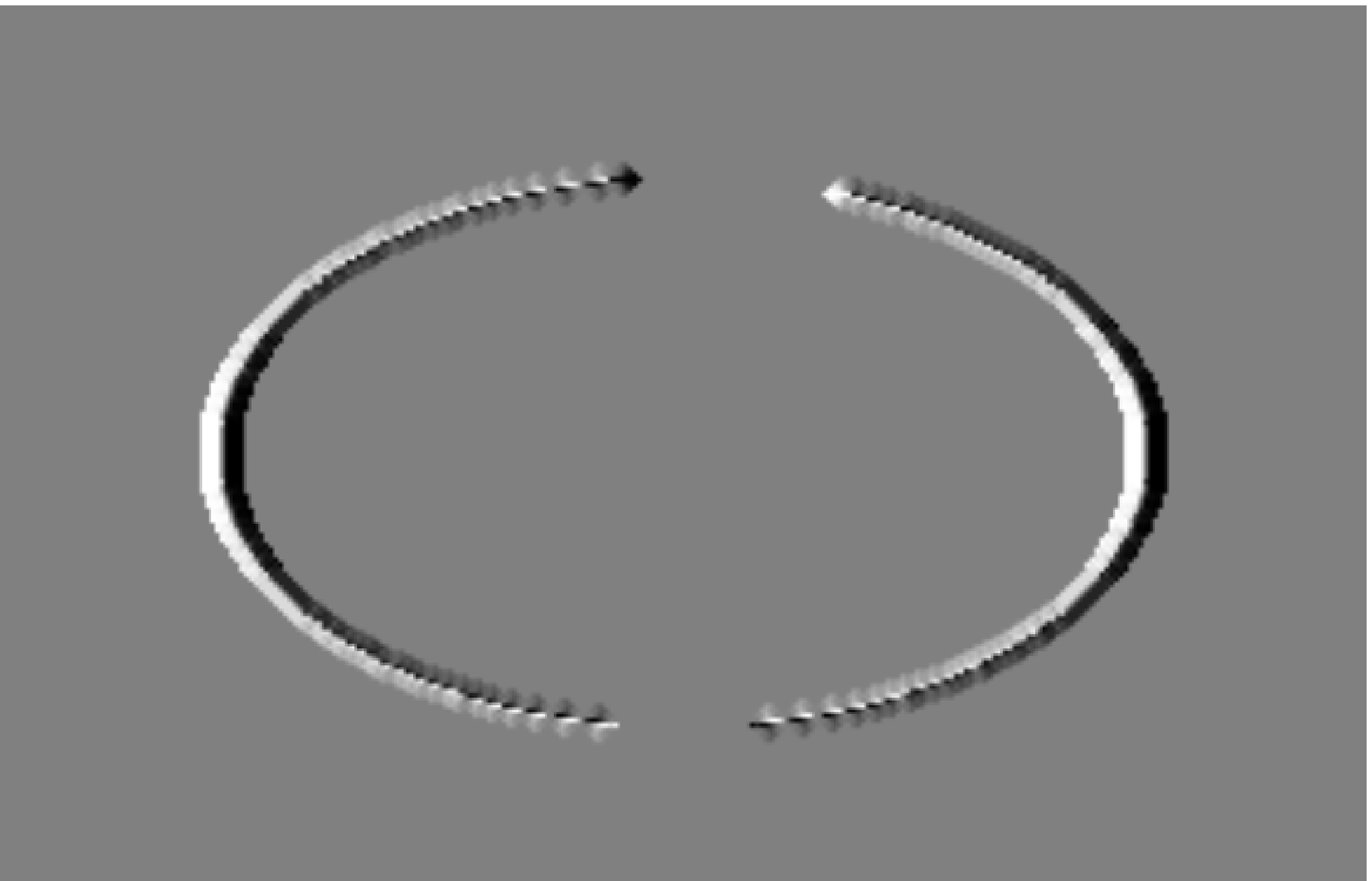,width=0.33\textwidth}
                \epsfig{figure=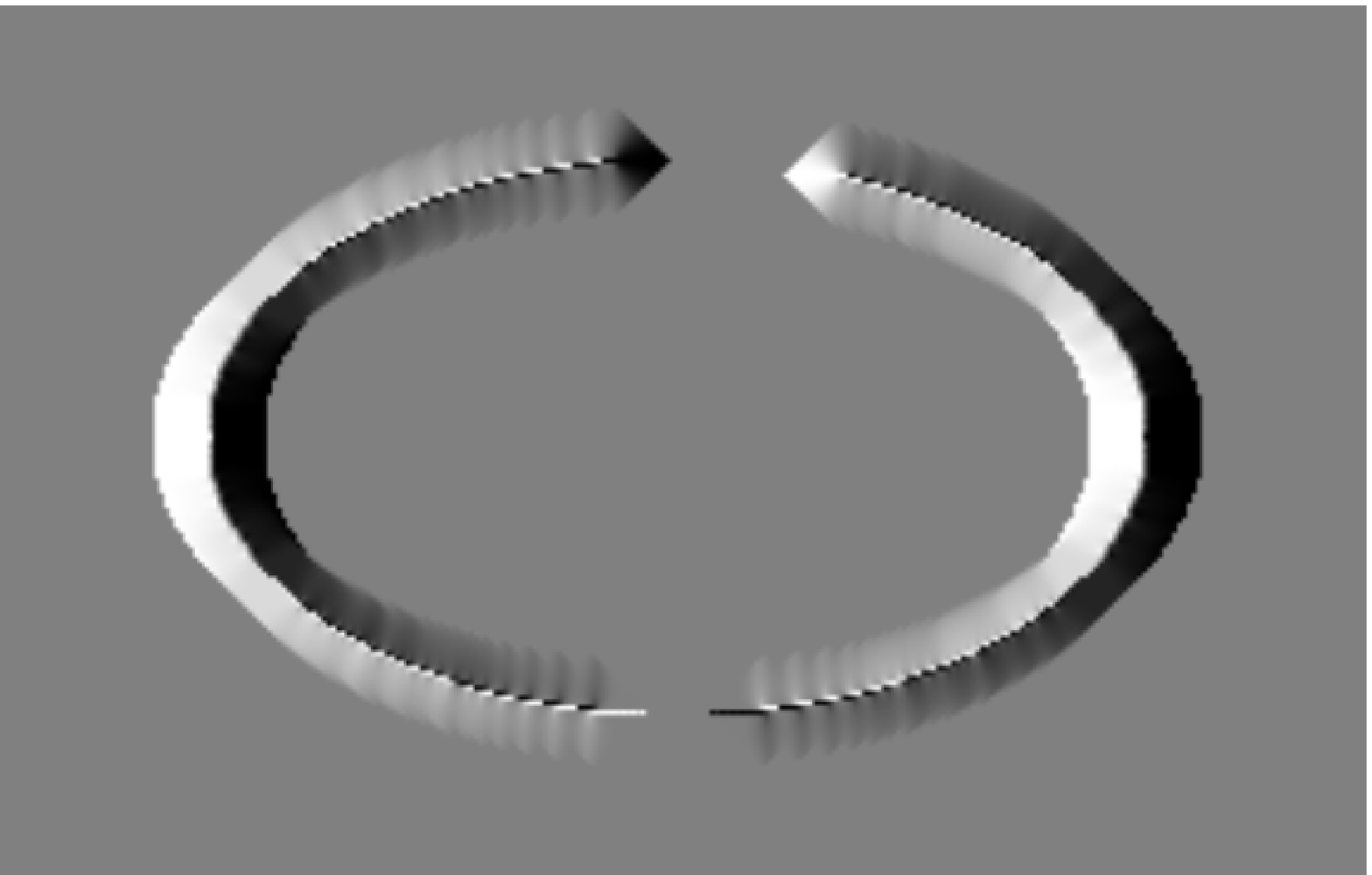,width=0.33\textwidth}}

    \vspace{2mm}
    \centerline{GVF y \hspace{3mm} 
		\epsfig{figure=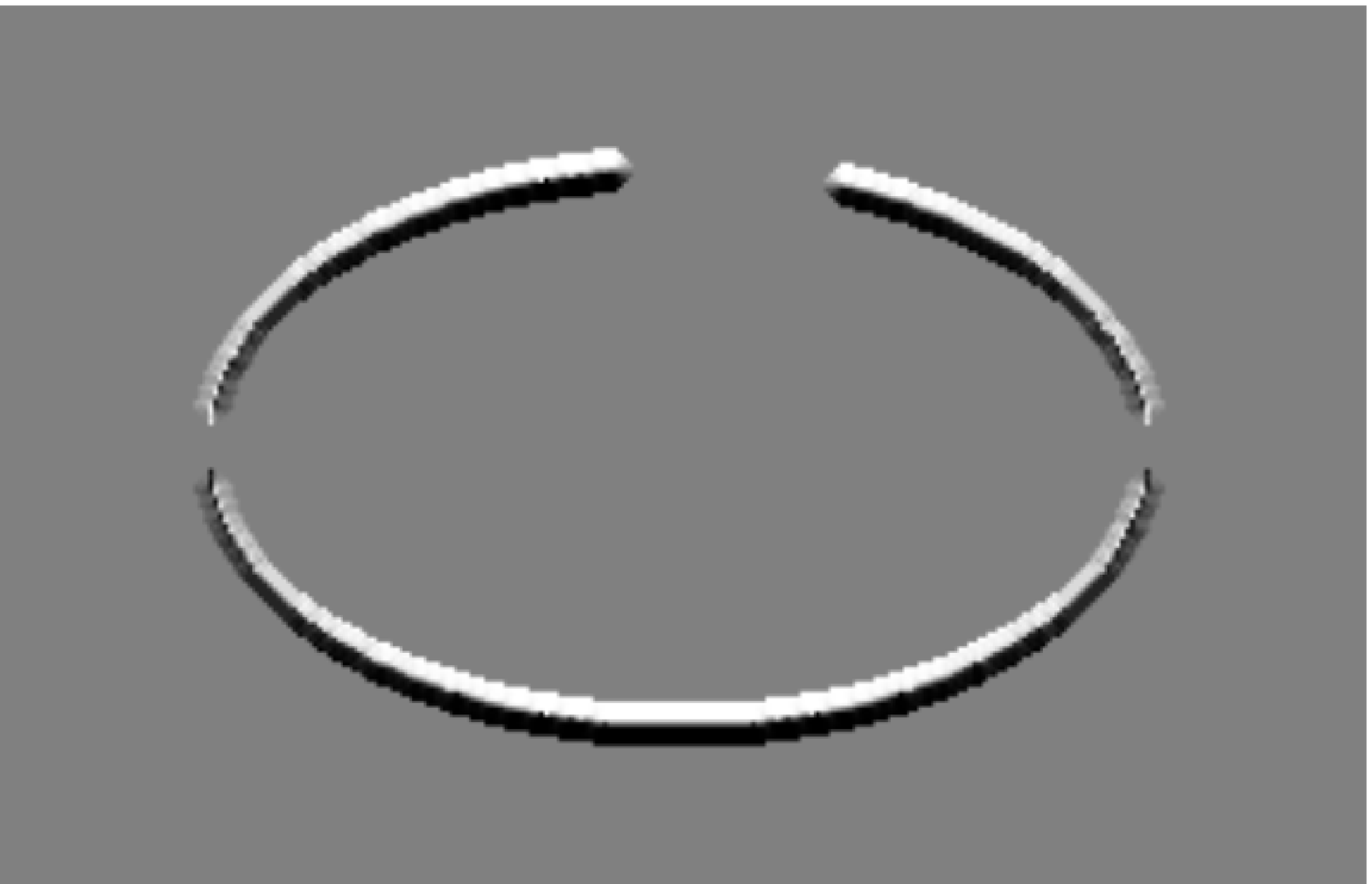,width=0.33\textwidth}
                \epsfig{figure=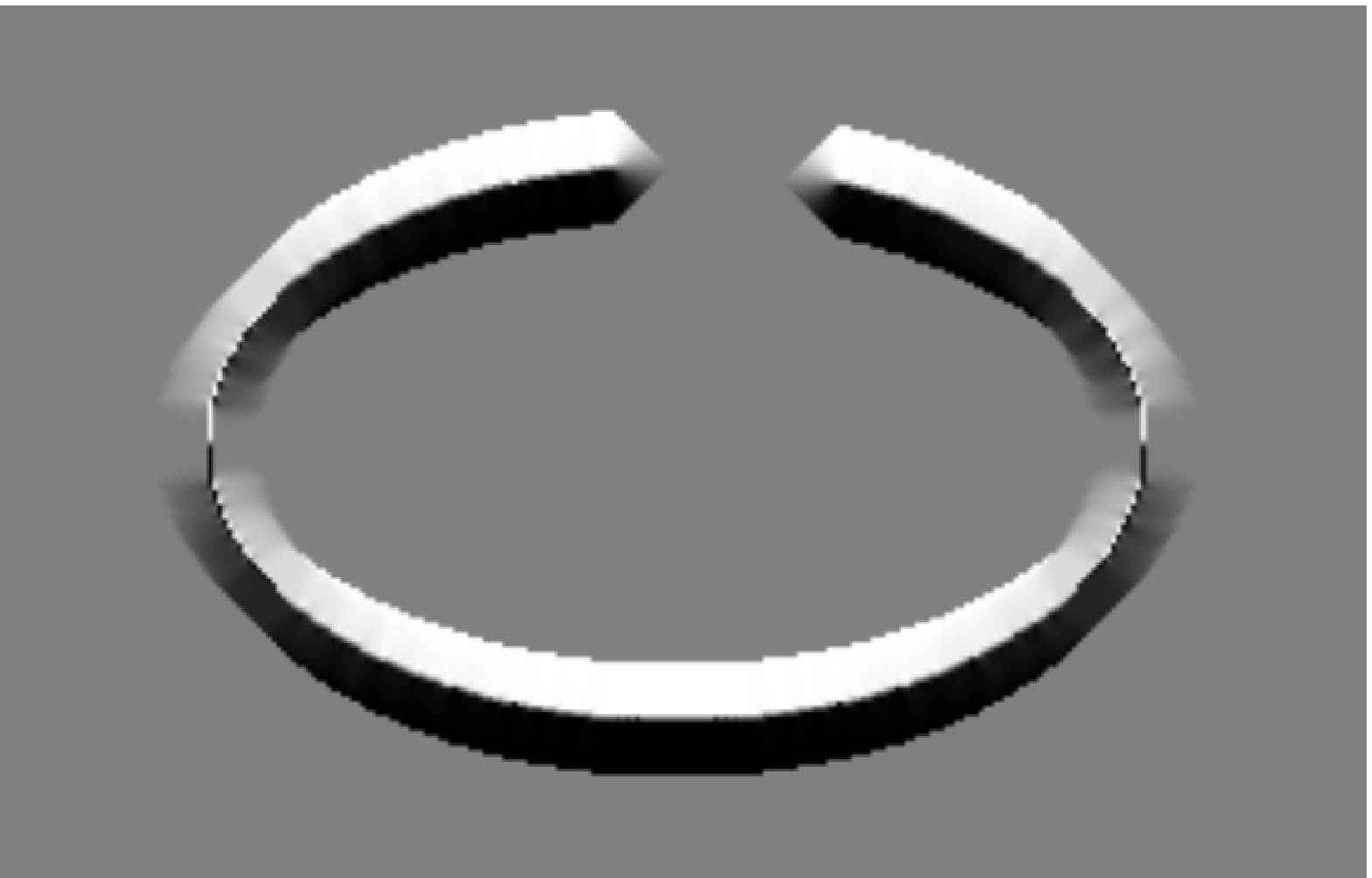,width=0.33\textwidth}}
                
    \caption{Gradient Vector Flow (GVF) in x and y directions, computed on the binary edge image with 3 and 10 iterations. In the GVF images, the values are normally in [-1,+1] range; pixel values are scaled to 0--255 range for displaying; light values are in $+x/+y$ direction (positive) and dark values are in $-x/-y$ direction (negative).}
    \label{fig:gvf}
\end{figure}

The edge term ($E_{edge}$) is calculated from the gray or binary edge map of the image by using the Gradient Vector Flow (GVF)~\cite{xu98} algorithm. Figure~\ref{fig:gvf} shows two examples of GVF computed on a binary edge image with 3 and 10 iterations. As the number of iterations increases, the capture range also increases. If the image is textured and contains edges that are close to each other, the number of iterations should be kept small in order to avoid jumping snakelets across the edges, especially if the stiffness is small.

The growing term ($E_{grow}$) is responsible for growing the snakelets. As shown in Figure~\ref{fig:snakelet}, growing forces act on the snakelet at one end or both ends; hence, the snakelet can grow in one direction (unidirectional) or in both directions (bidirectional). The magnitude of this force is proportional to the gradient magnitude in grayscale edge magnitudes, and it is taken as constant in binary edge images.

\section{Edge Recovery with Snakelets}

In edge recovery, we only have an edge image as input (original image or gradients are not available or not used). The edges may have breaks and our goal is to recover the broken edges using only the input edge image.
We depict the edge recovery algorithm with a simplified example, shown in Figure~\ref{fig:ellipse}, where there is an ellipse with a break in the binary image. The recovery algorithm works as follows:

\begin{figure}[h!t]

    \vspace{2mm}
    \centerline{Snakelet initial \hspace{35mm} GVF x \hspace{40mm}  GVF y \hspace{10mm}}
    \centerline{\epsfig{figure=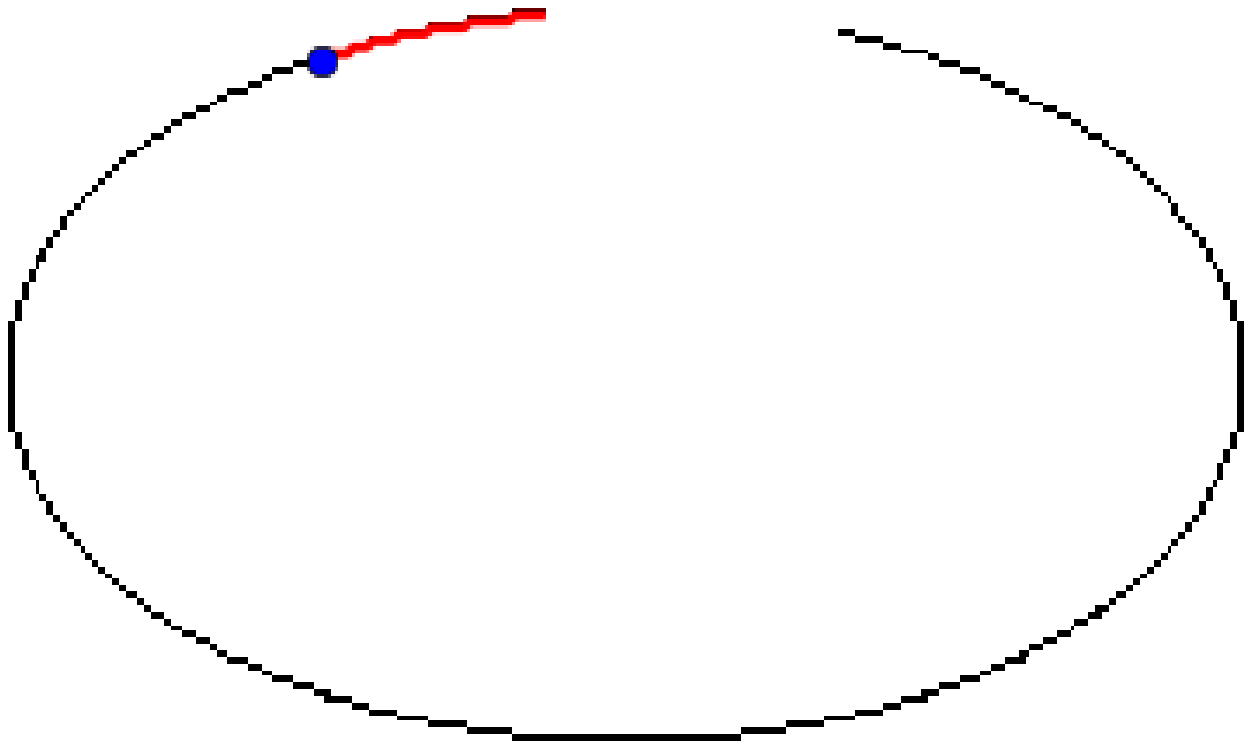,width=0.33\textwidth} 
		\epsfig{figure=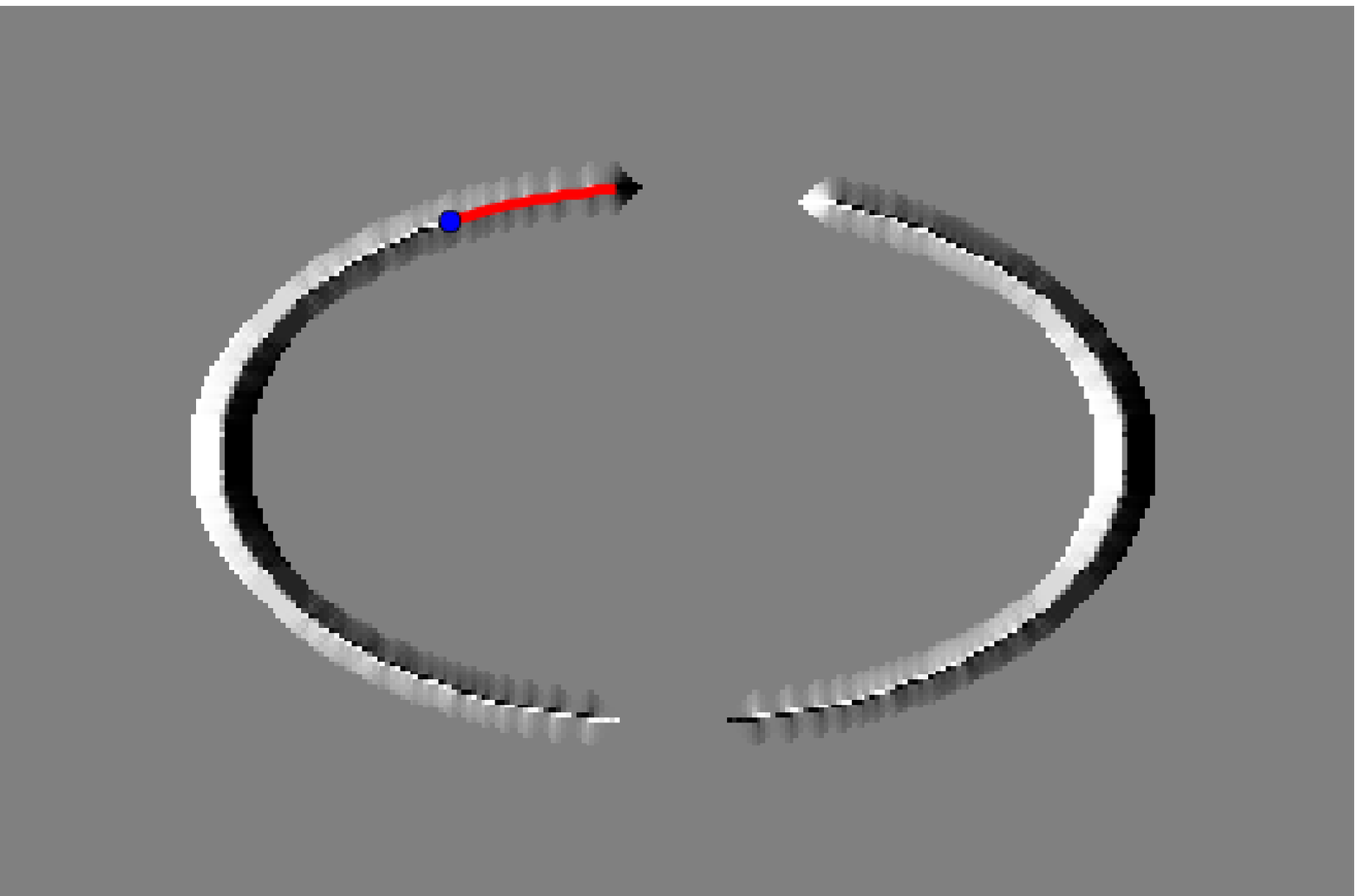,width=0.33\textwidth}
                \epsfig{figure=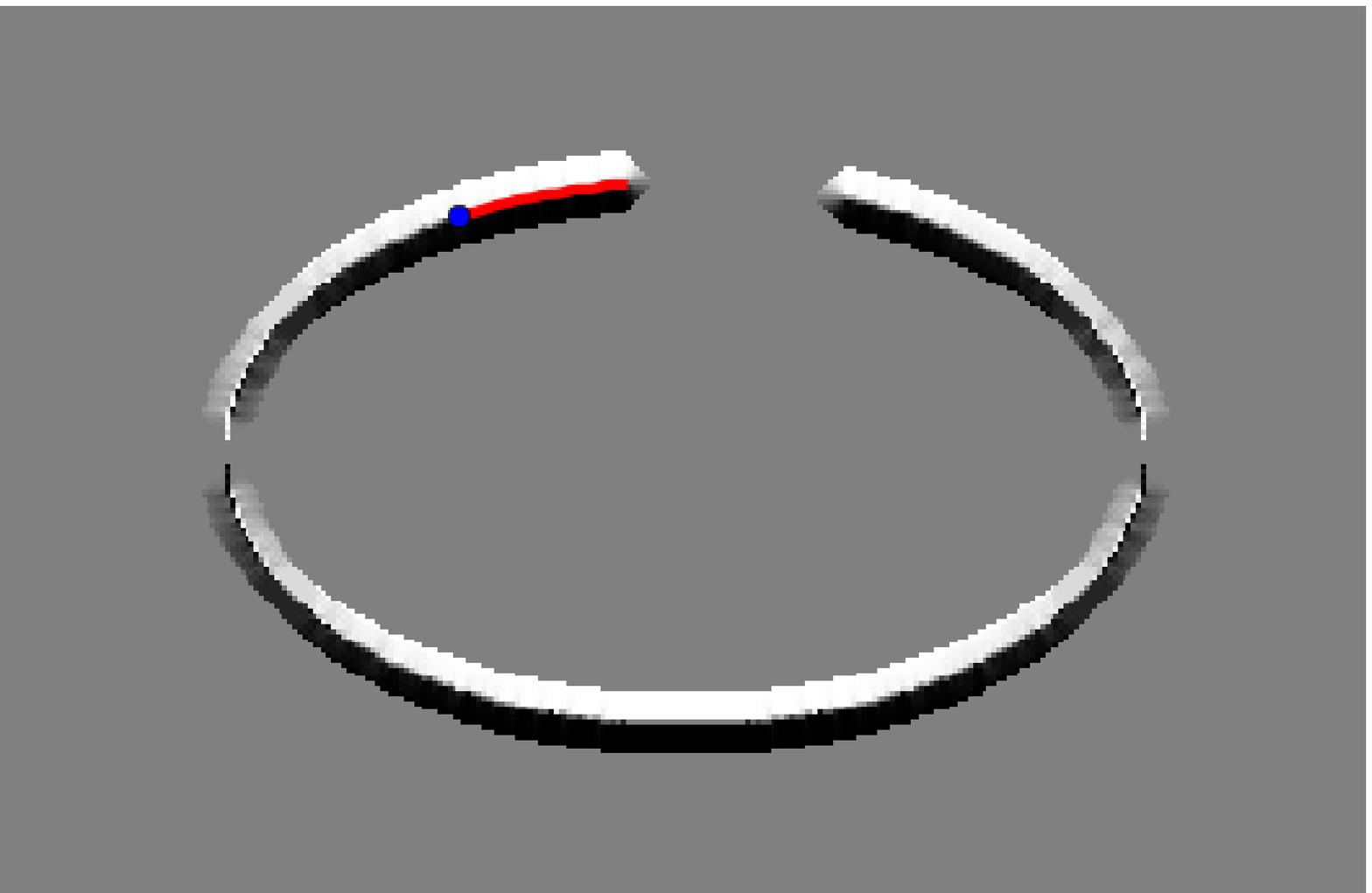,width=0.33\textwidth}}
    \vspace{2mm}
    \centerline{Snakelet final \hspace{100mm}}
    \centerline{\epsfig{figure=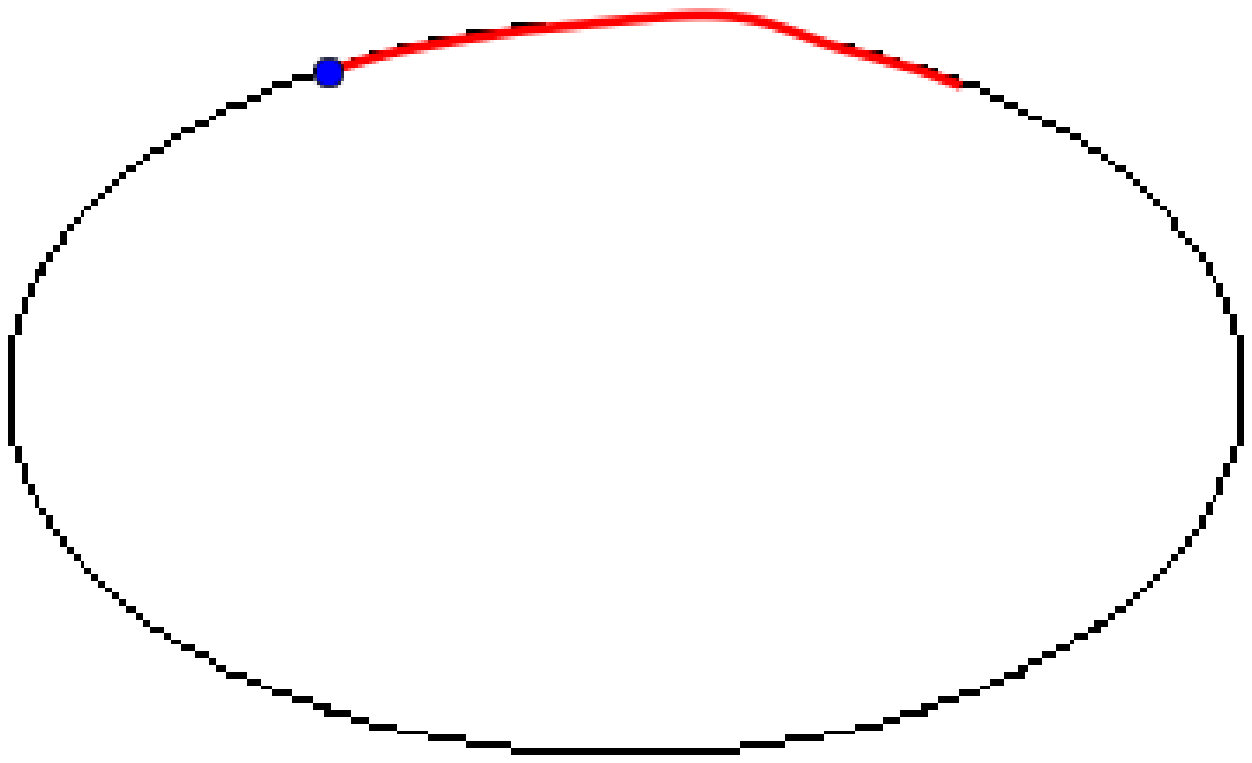,width=0.33\textwidth} 
		\epsfig{figure=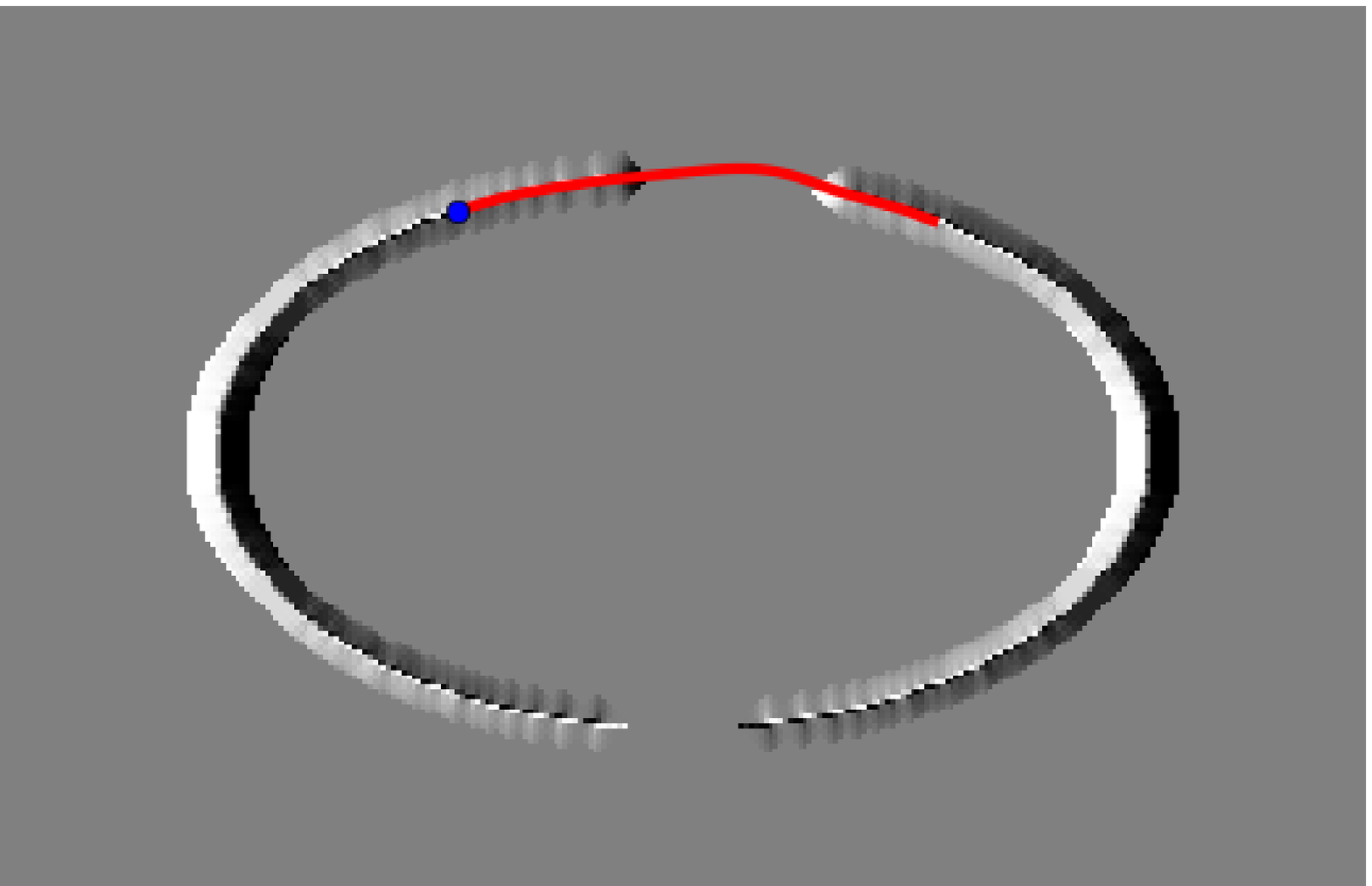,width=0.33\textwidth}
                \epsfig{figure=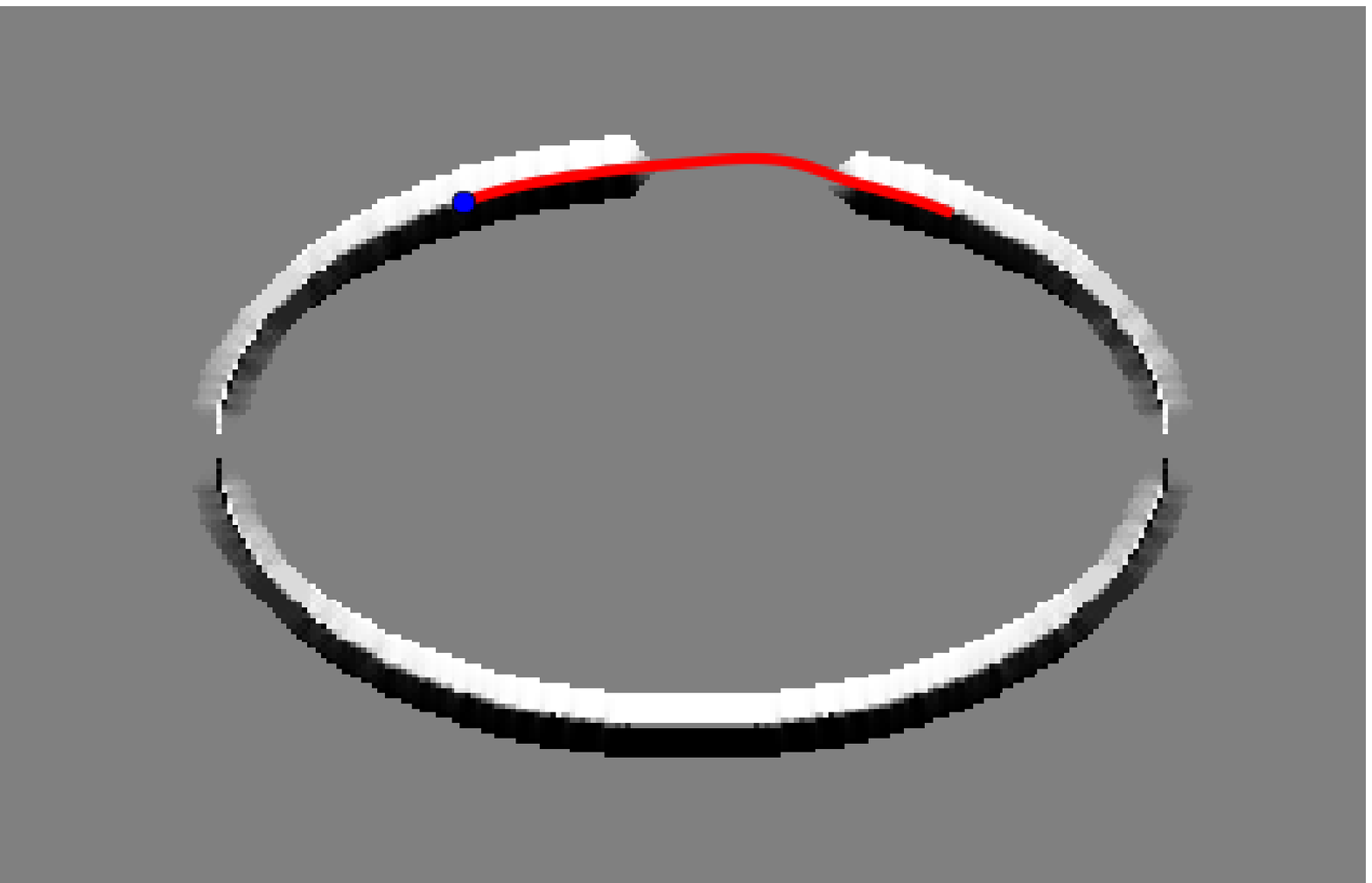,width=0.33\textwidth}}
    \caption{Edge recovery. Evolution of a single snakelet under the effect of gradient vector flow (GVF) in x and y directions, computed on the binary edge image with 5 iterations. The filled blue circle is the start of the snakelet, which grows in the other end to reach the other side of the break.}
    \label{fig:ellipse}
\end{figure}

\begin{itemize}
 \item \textit{Initialization.} We initialize snakelets at the end points of breaks, starting from the endpoint of the edge and tracing backwards. In Figure~\ref{fig:ellipse}, the ellipse has a single break and two endpoints; hence we initialize two snakelets at each end point. In the example, we show only one of the snakelets for clarity. The initial length of the snakelet should be so long as to capture the local edge shape. If it is too short, it will behave like a straight line, if it is too long it is computationally expensive. We normally use a length of $20-30$ pixels for the initial snakelets, if the edge fragment is long enough; otherwise it is as long as the edge fragment. There must be at least 2 points to initialize a snakelet.
 
 \item \textit{GVF.} Initially the GVF is computed over the edge image with a few iterations. It is iteratively expanded until a solution is found or a maximum number of iterations is reached.

 \item \textit{Deformation and Growing.} The snakelets deform under the effect of external force GVF and internal forces (tension and stiffness) and grow alternately. The internal tension and GVF at the end of an edge contracts the snakelet, while the growing force elongates it. The growing force should be larger than the contraction forces so that the snakelet can grow. The snakelet grows at one end until its length reaches a predefined maximum length. Then, if it reached an edge, the process completes; otherwise, GVF is expanded and the whole process is repeated starting with the initial snakelet. If a snakelet fails to reach an edge, it is discarded. Breaks larger than the maximum growing length cannot be recovered.
 
 This iterative GVF expansion and deformation and growing continue until a maximum number of GVF iterations is reached. The idea is that if the end points of the breaks are aligned, then a small number of GVF iterations is sufficient to find a solution; otherwise, a larger number of GVF iterations will be needed to increase the capture range and attract the snakelet to the other side of the broken edge. Starting with a large number of GVF iterations causes the snakelets to jump across edges that are close to each other; this is not desirable.
 
 Figure~\ref{fig:ellipse} shows a single snakelet initialized at the end of a broken edge and deformed and grew to find the other end of the edge to recover the break. The initial length was $35$ and maximum growing length was $70$. The initial number of GVF iterations was 5 and it was sufficient to find a solution, expanding the GVF was not needed.
 
\end{itemize}

\begin{figure}[h!t]
    \centerline{Input edge image \hspace{15mm} Initial snakelets \hspace{15mm}  Final snakelets}
    \centerline{\epsfig{figure=figures/U1-bow.eps,width=0.34\textwidth} 
		\epsfig{figure=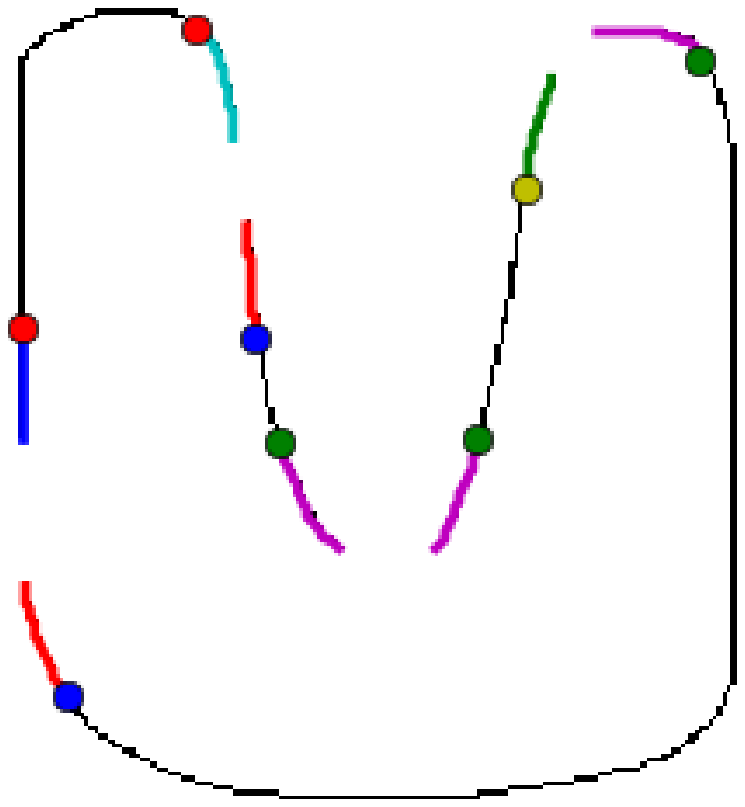,width=0.30\textwidth}
                \epsfig{figure=figures/U1-wob-final-snakes.eps,width=0.30\textwidth}}
		
    \centerline{\epsfig{figure=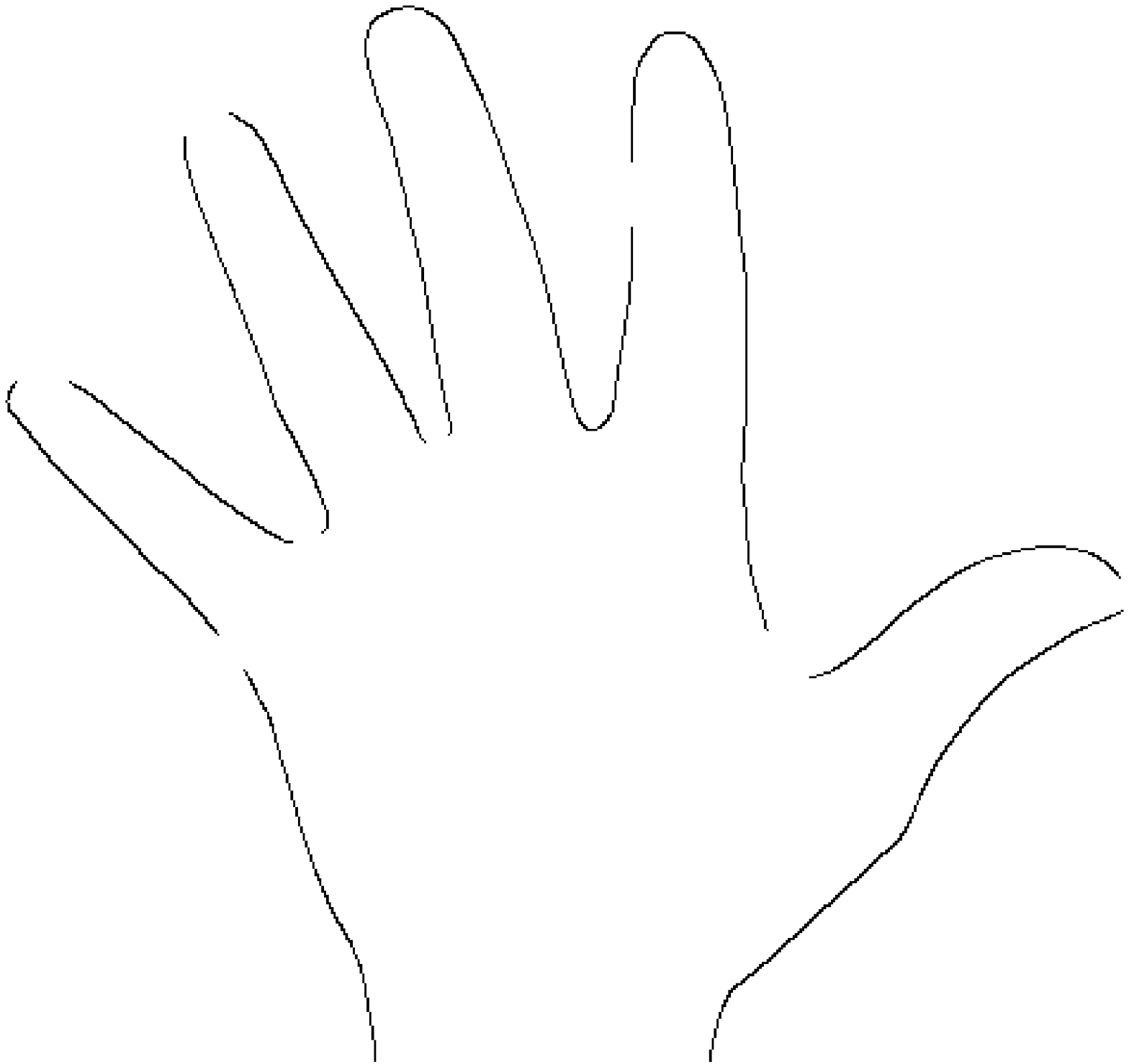,width=0.34\textwidth} 
		\epsfig{figure=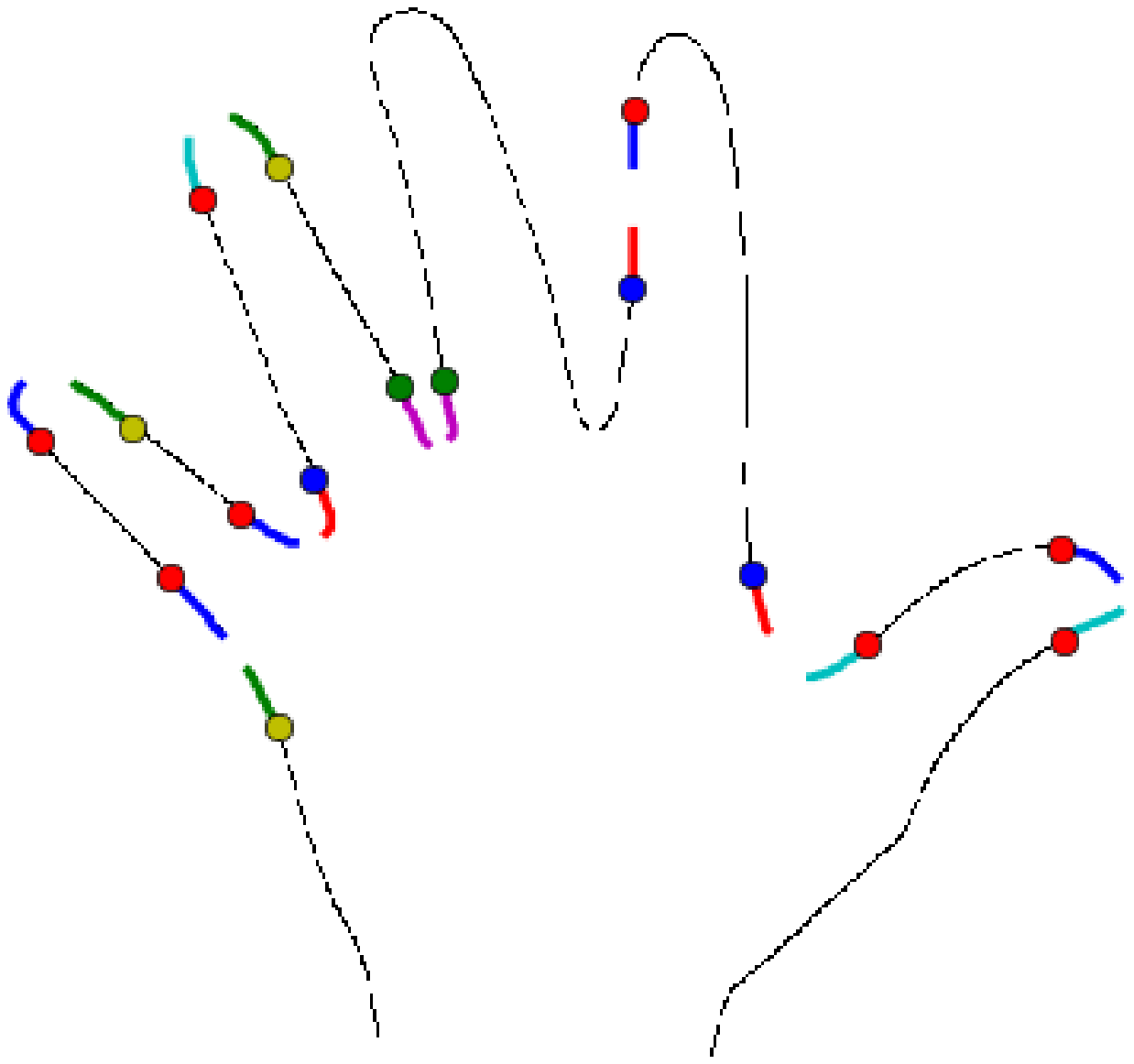,width=0.32\textwidth}
                \epsfig{figure=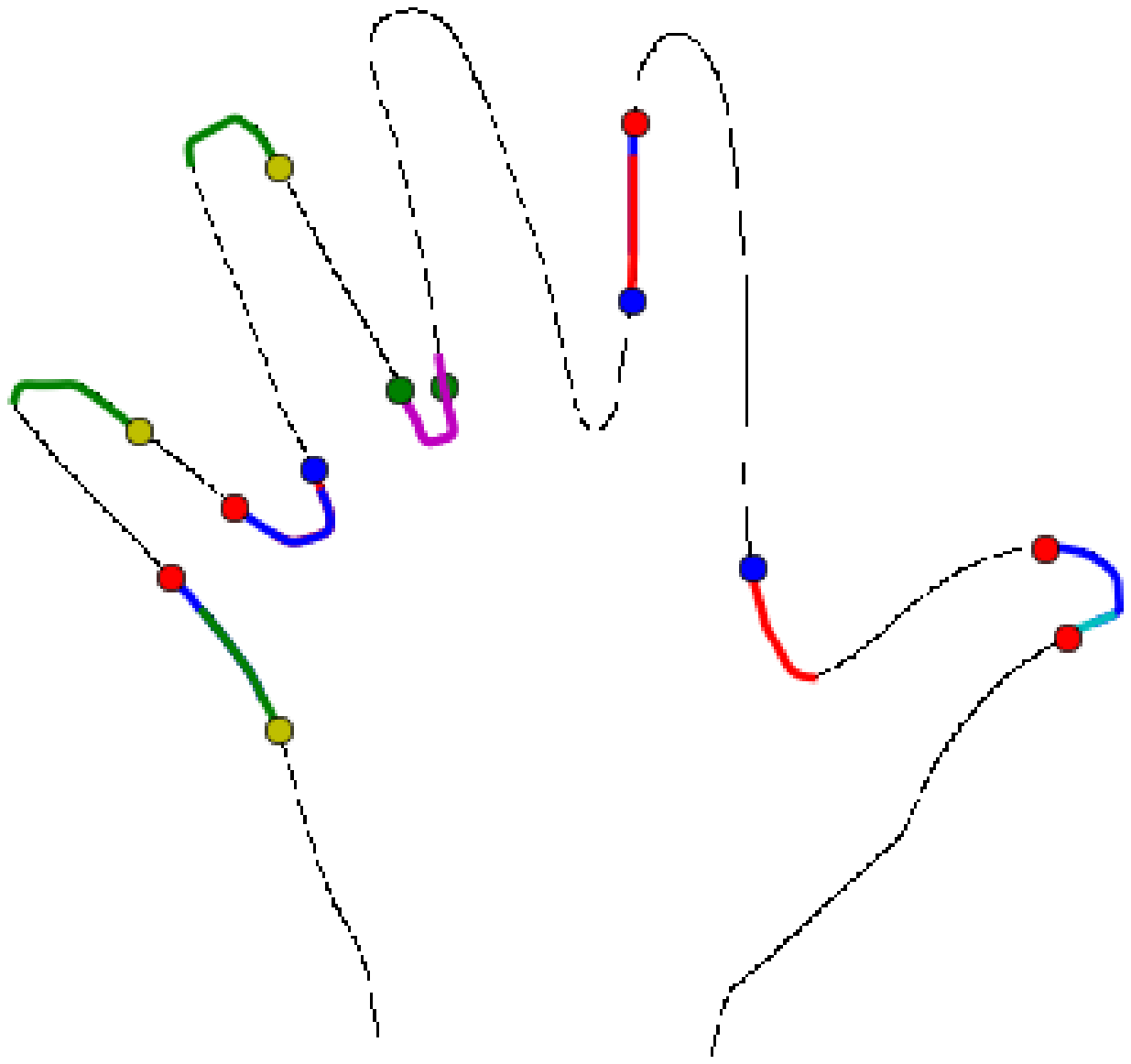,width=0.32\textwidth}}

     \centerline{\epsfig{figure=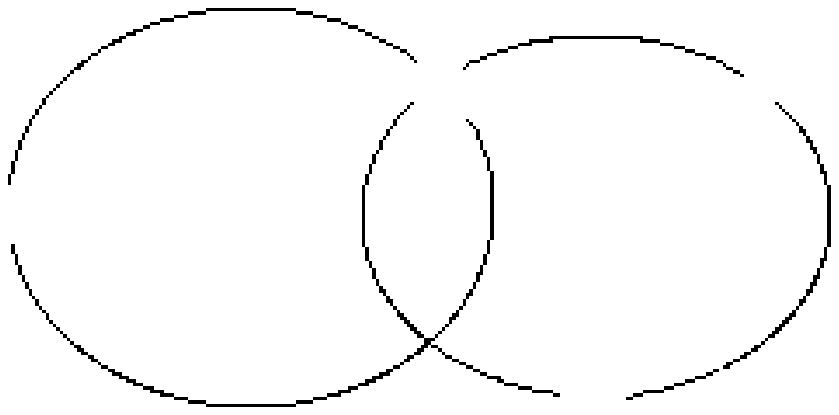,width=0.35\textwidth} 
		\epsfig{figure=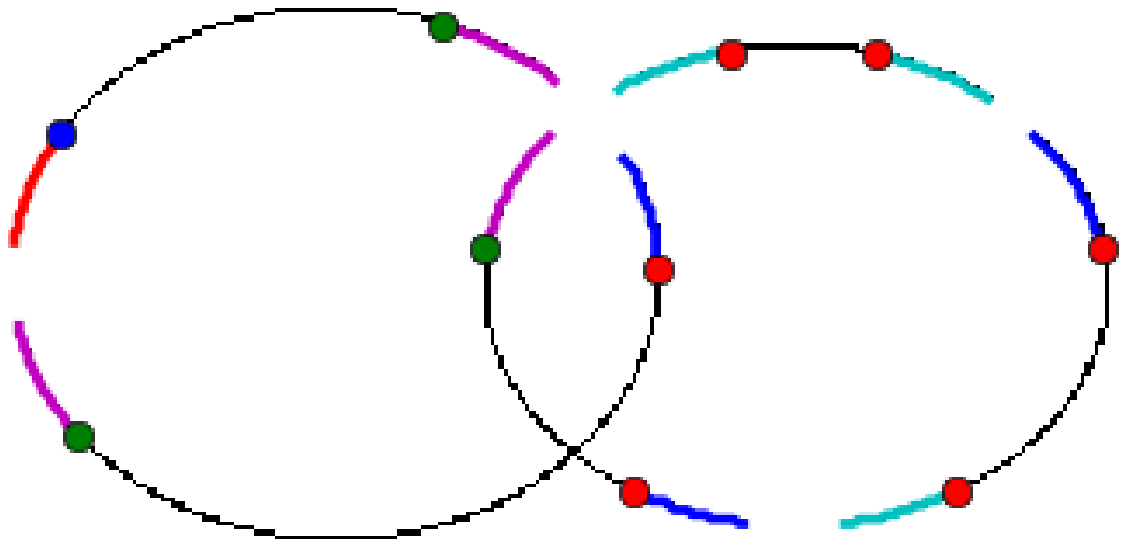,width=0.31\textwidth}
                \epsfig{figure=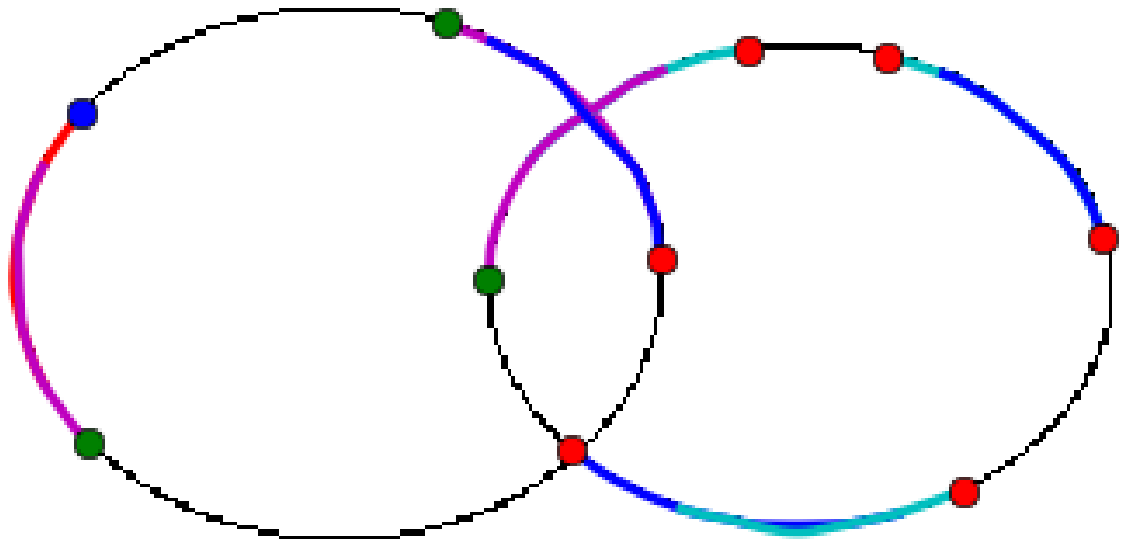,width=0.31\textwidth}}

     \centerline{\epsfig{figure=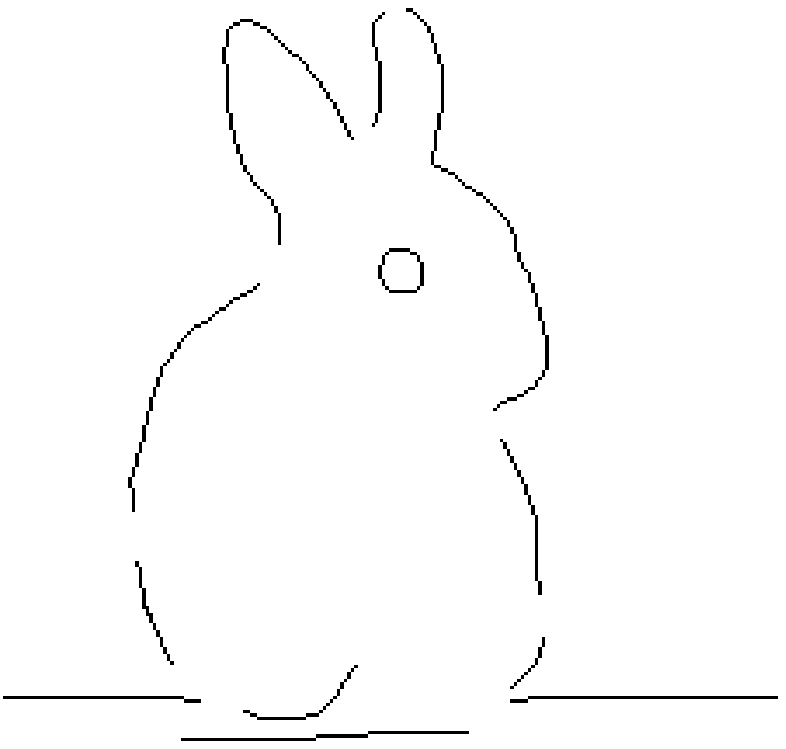,width=0.33\textwidth} 
		\epsfig{figure=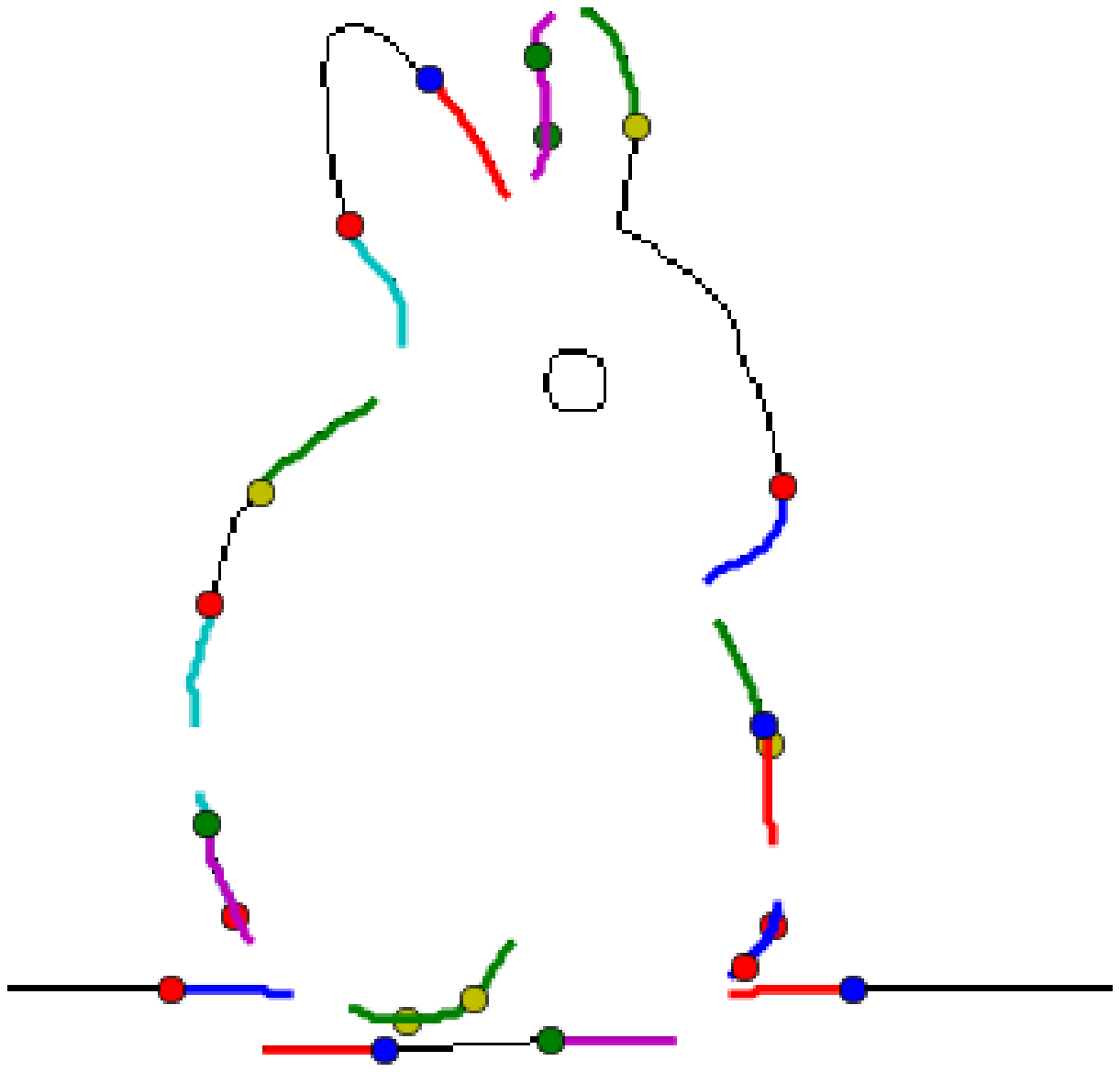,width=0.33\textwidth}
                \epsfig{figure=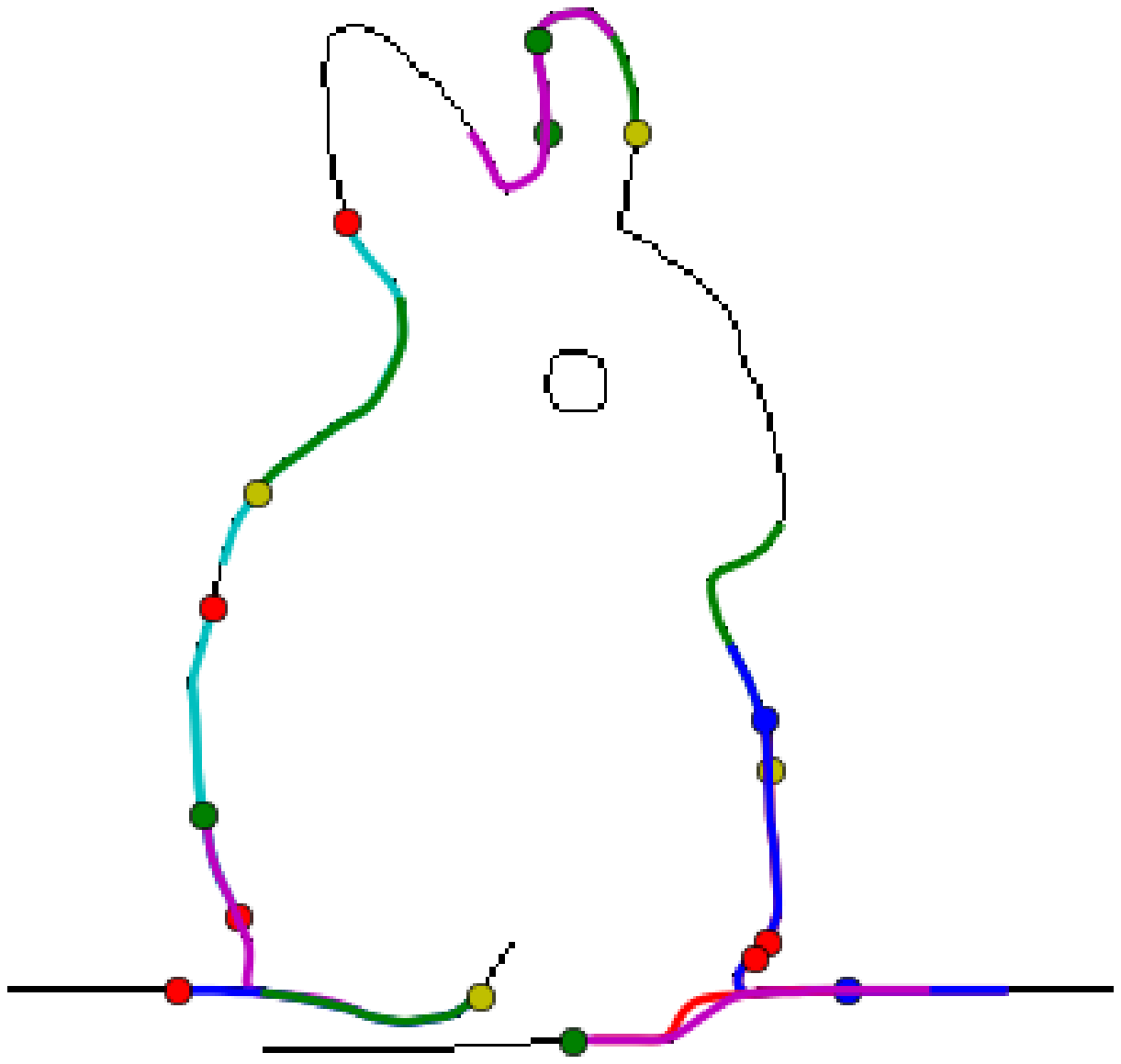,width=0.33\textwidth}}
                
    \caption{Recovering broken edges. Left: input edge image, middle: initial snakelets, right: final snakelets. Filled circles are starting point of the unidirectional snakelets.}
    \label{fig:snake-recover}
\end{figure}

Figure~\ref{fig:snake-recover} shows examples of edge recovery. The inputs are binary edge images with artificially generated breaks of different lengths (left images). Snakelets are initialized at the end points of the breaks (middle images) and the above recovery algorithm is run. The final snakelets are shown on the right. If the end points of the breaks are aligned or converging or if the breaks are short, they are easily recovered. The difficult cases are when the end points of the break are not aligned or not converging, as in the middle break of the U shape and finger tips in the hand image. The algorithm tries to recover such difficult breaks by iteratively expanding the GVF to increase the capture range and pull the snakelet towards the other end of the break. Even then, a solution may not be found, as in the finger tips of the hand image, where only one of the pair of snakelets was able to find a solution and the others were discarded.

\begin{figure}[h!t]
    \centerline{Input edge image \hspace{20mm} Initial snakelets \hspace{15mm}  Final snakelets}
    \centerline{\epsfig{figure=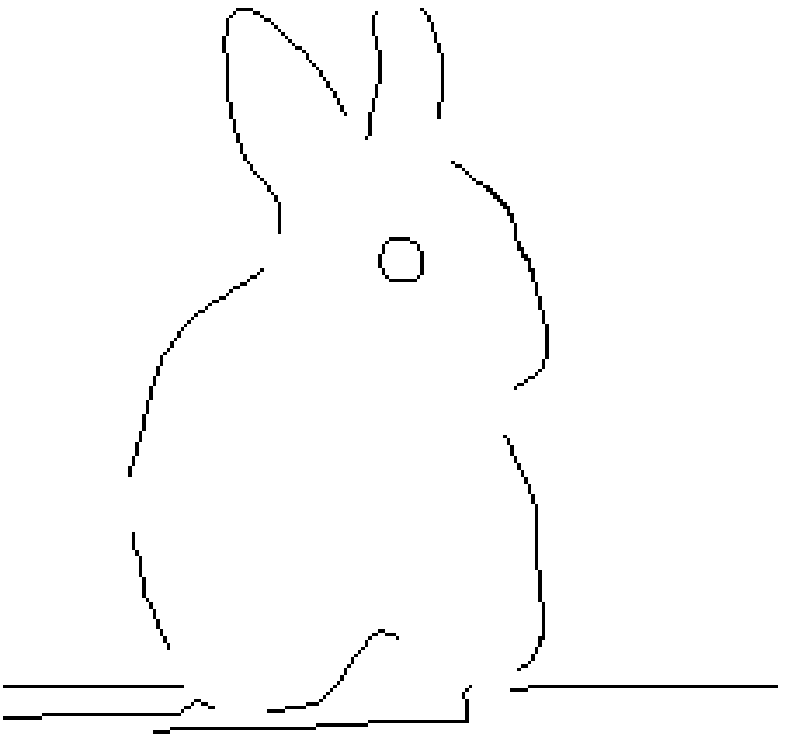,width=0.33\textwidth} 
		\epsfig{figure=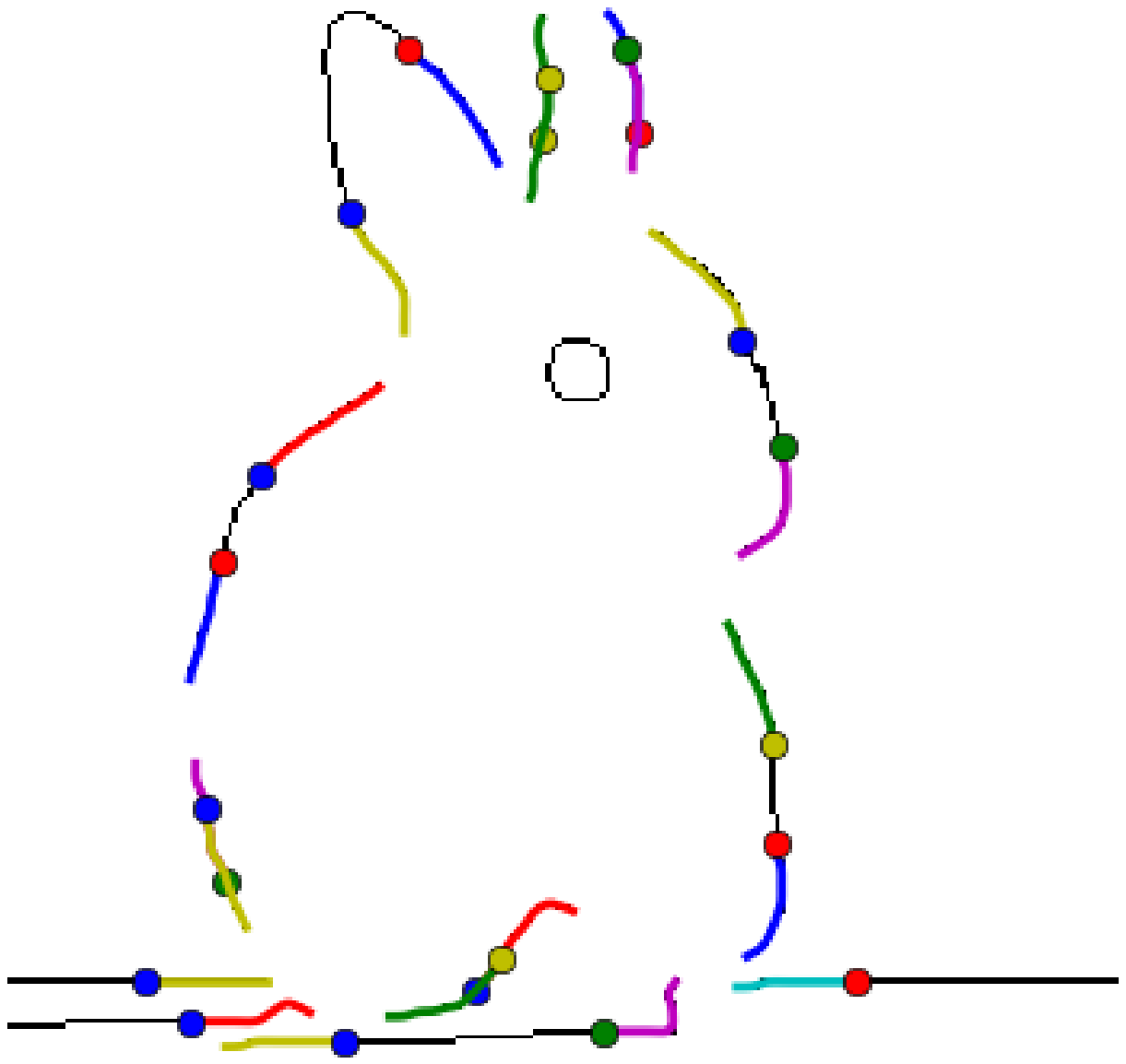,width=0.33\textwidth}
                \epsfig{figure=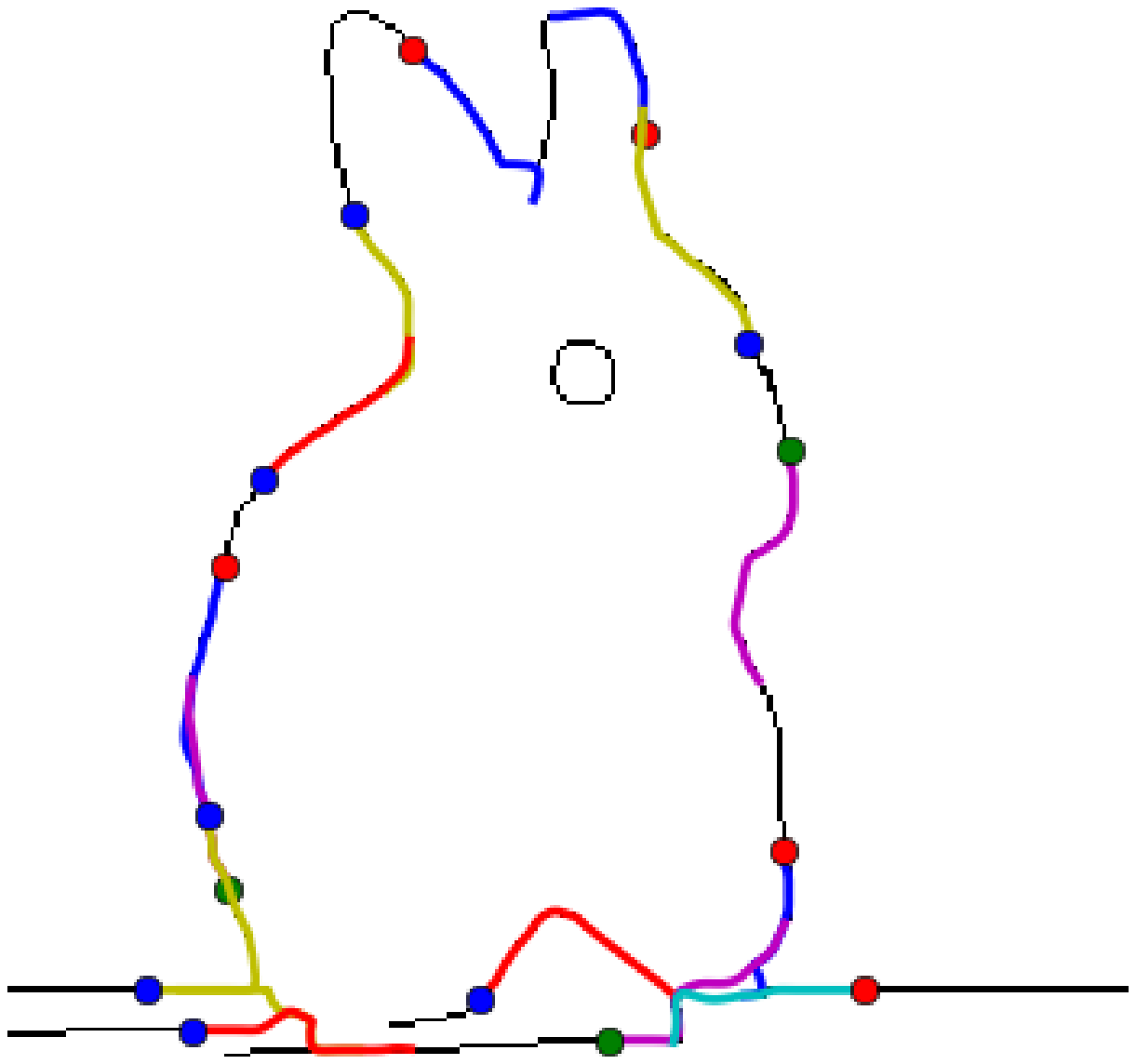,width=0.33\textwidth}}
    \vspace{2mm}
    \centerline{Input image \hspace{15mm} Initial snakelets + gradients \hspace{10mm}  Final snakelets}
    \centerline{\epsfig{figure=figures/bunny.eps,width=0.33\textwidth} 
		\epsfig{figure=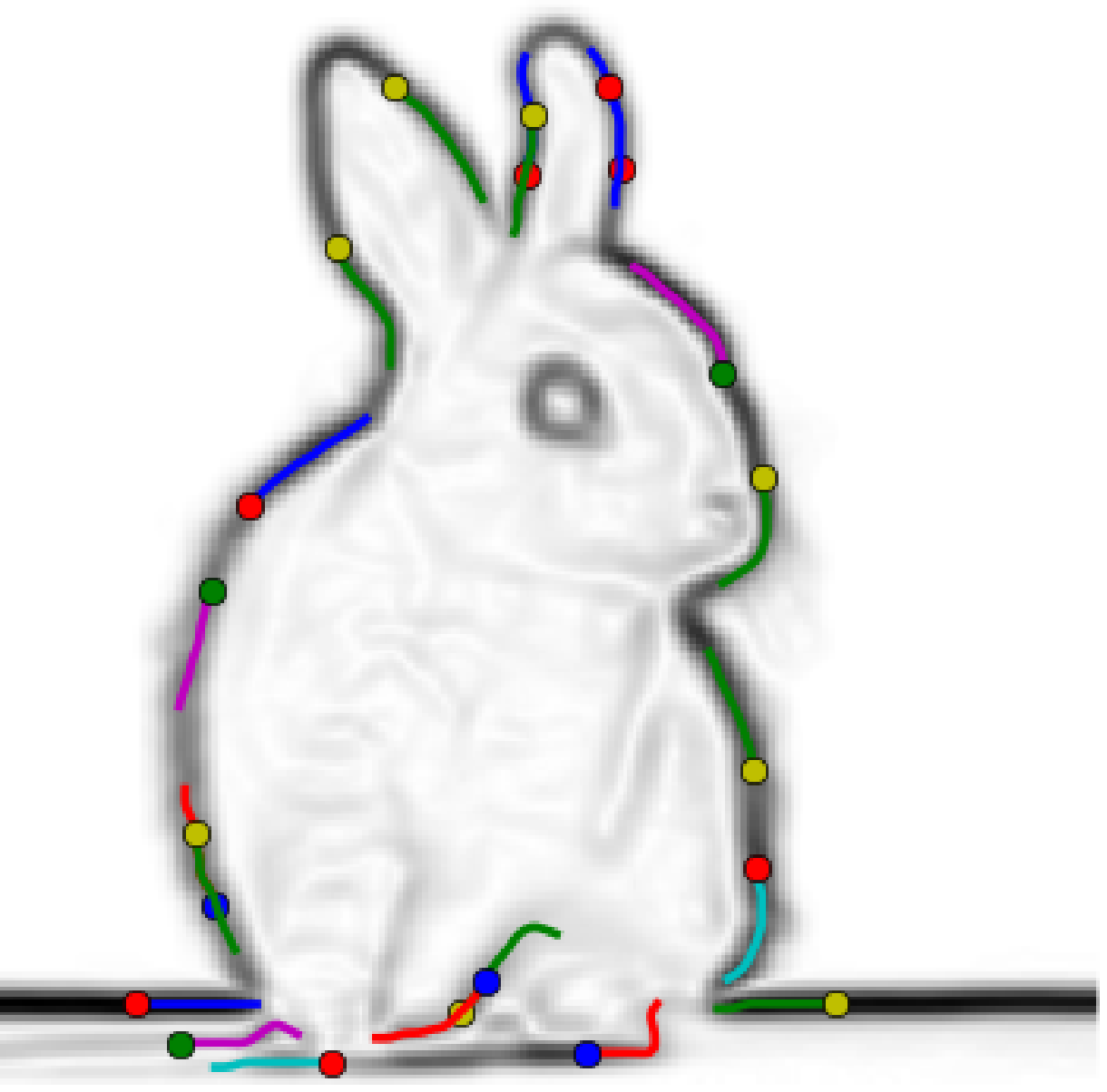,width=0.33\textwidth}
                \epsfig{figure=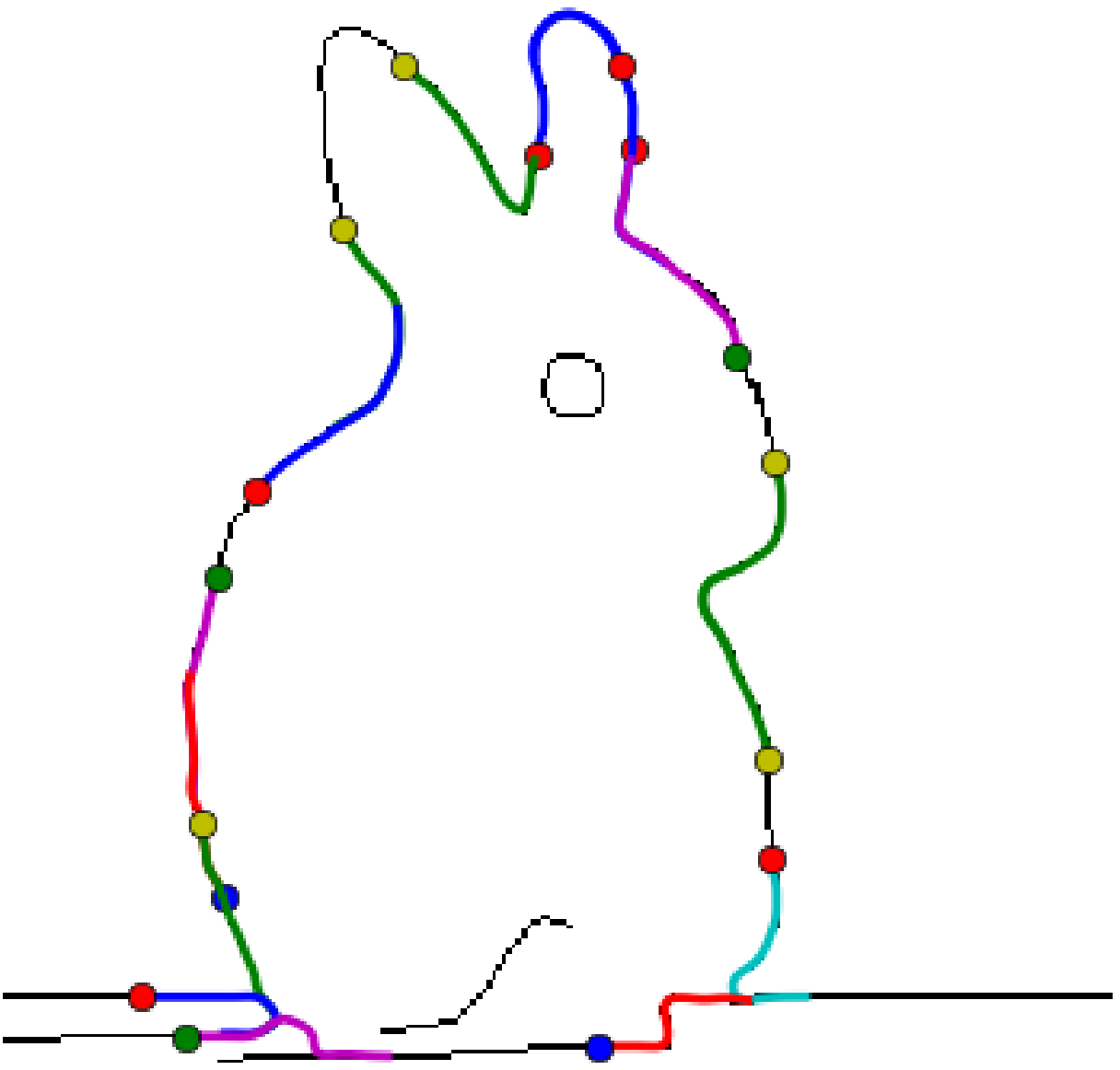,width=0.33\textwidth}}
    \vspace{5mm}
    \centerline{Input edge image \hspace{20mm} Initial snakelets \hspace{15mm}  Final snakelets}
    \centerline{\epsfig{figure=figures/hand1-bow.eps,width=0.33\textwidth} 
		\epsfig{figure=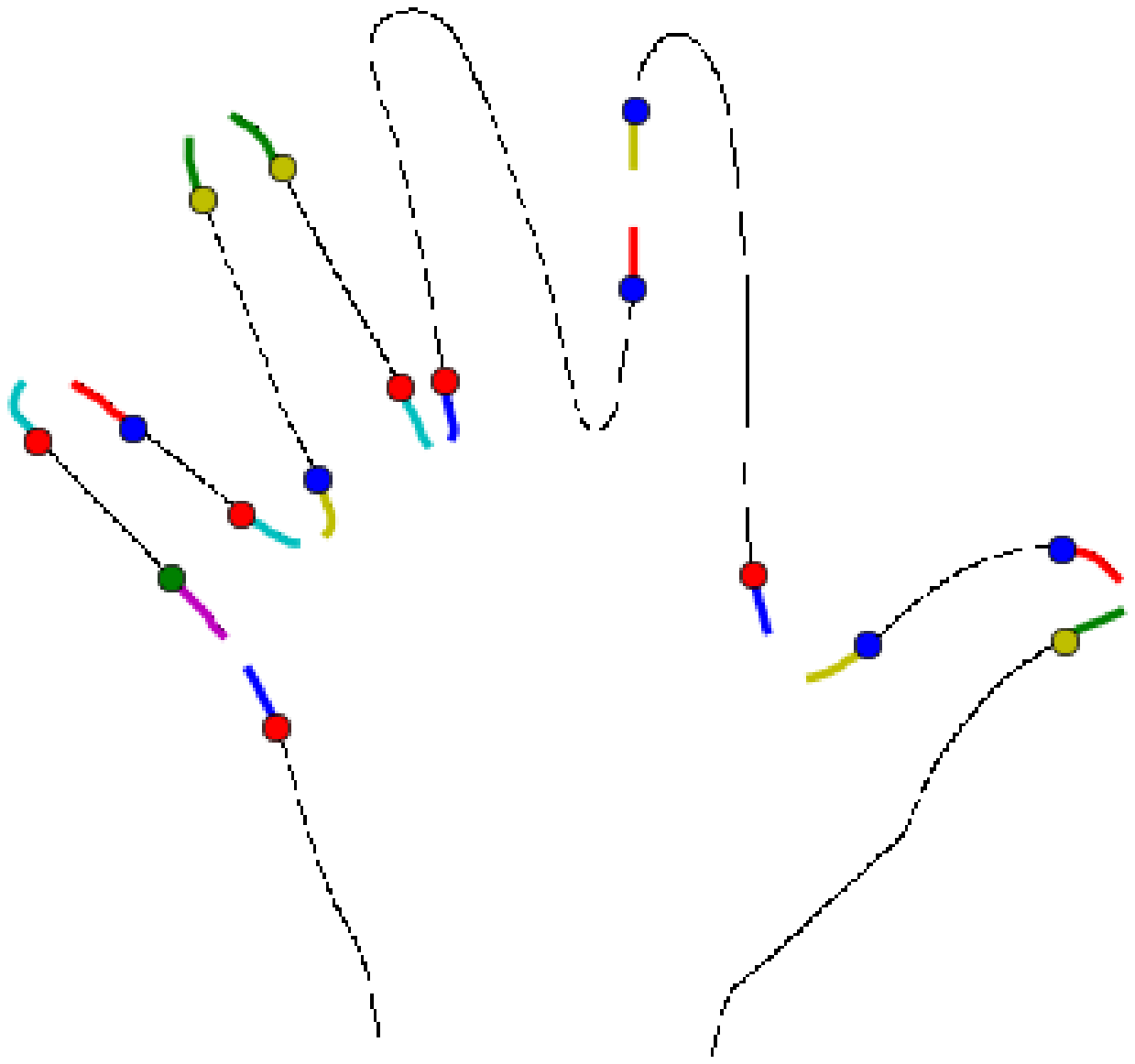,width=0.33\textwidth}
                \epsfig{figure=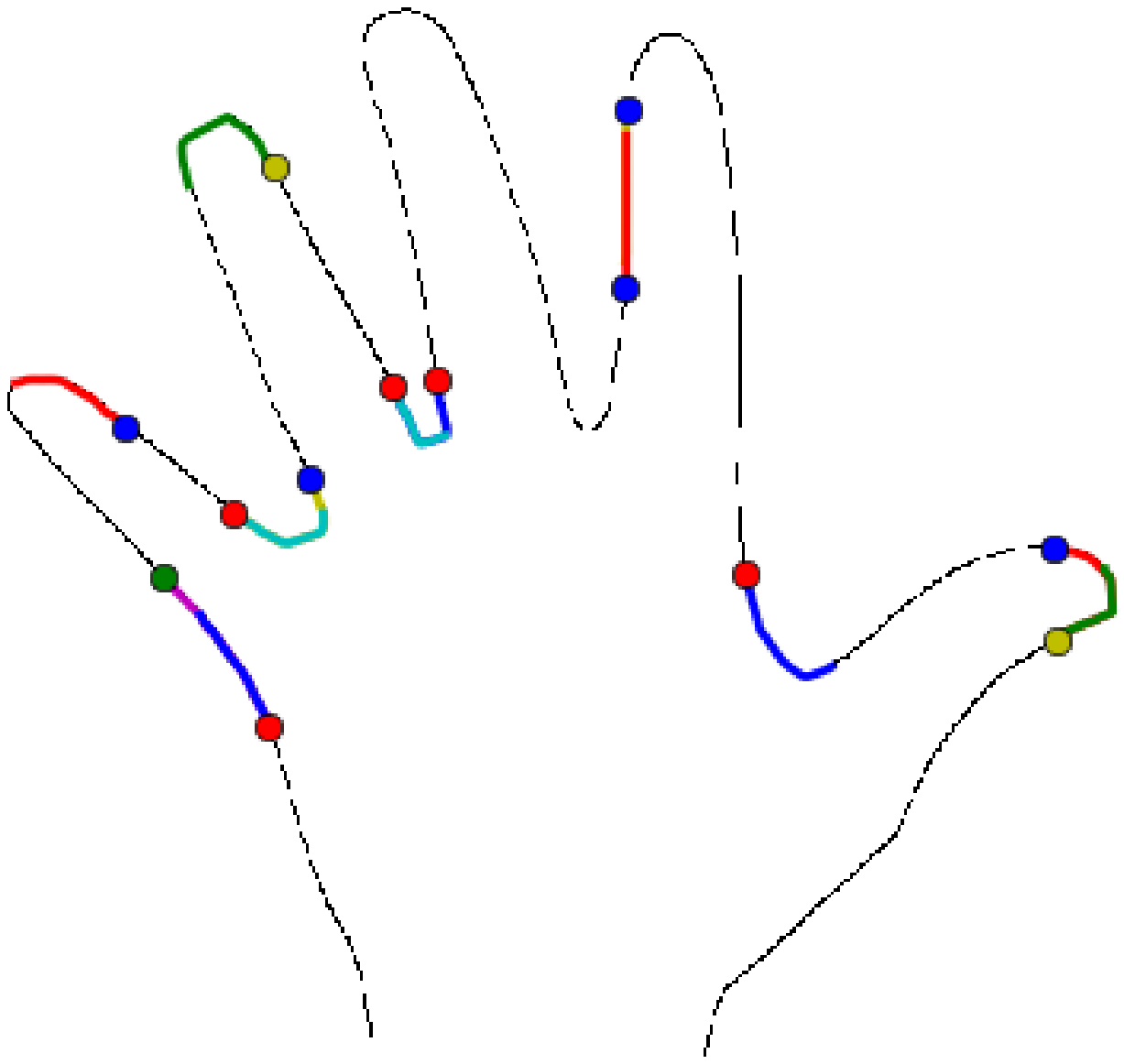,width=0.33\textwidth}}
    \vspace{2mm}
    \centerline{Input image \hspace{15mm} Initial snakelets + gradients \hspace{10mm}  Final snakelets}
    \centerline{\epsfig{figure=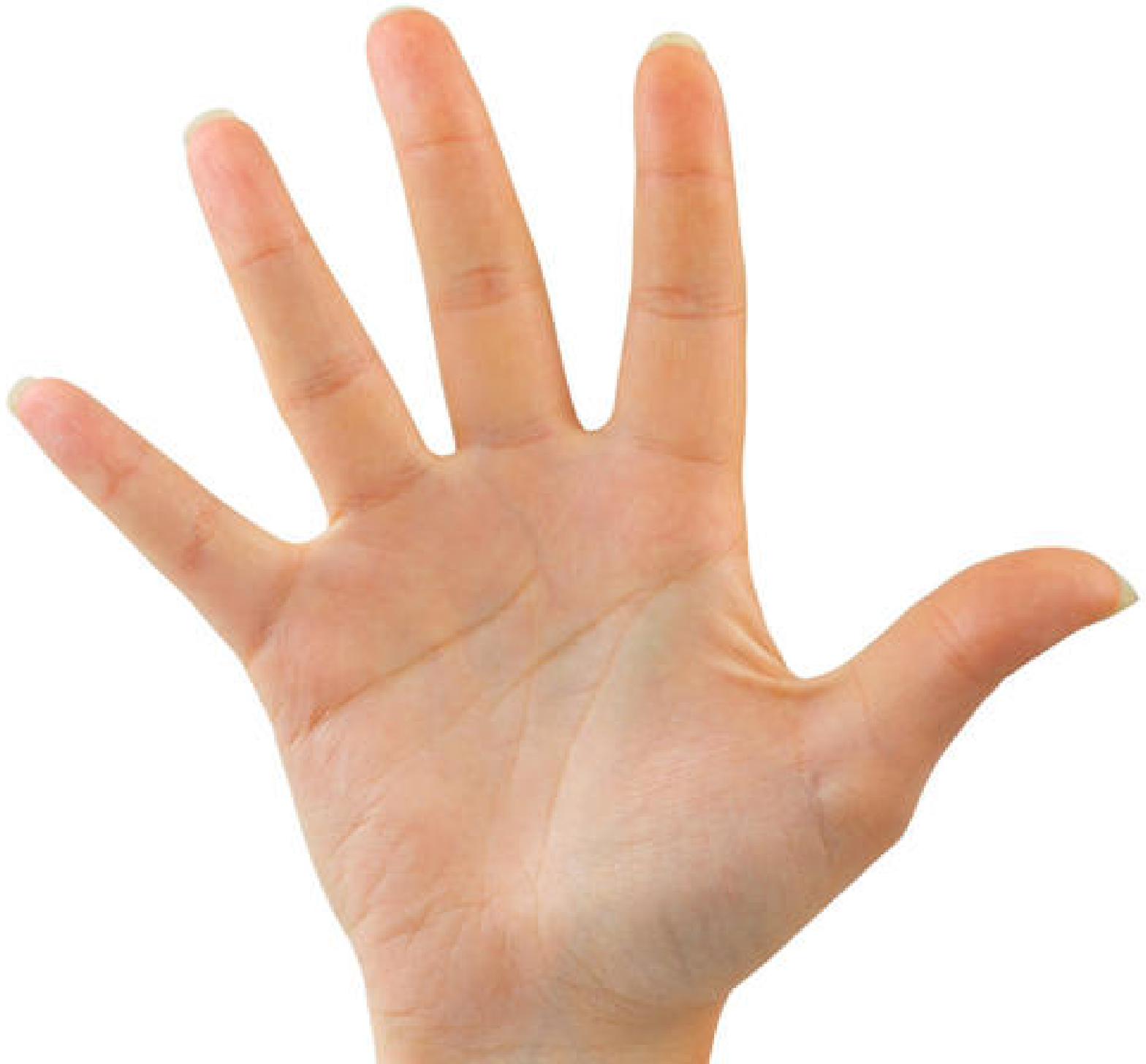,width=0.33\textwidth} 
		\epsfig{figure=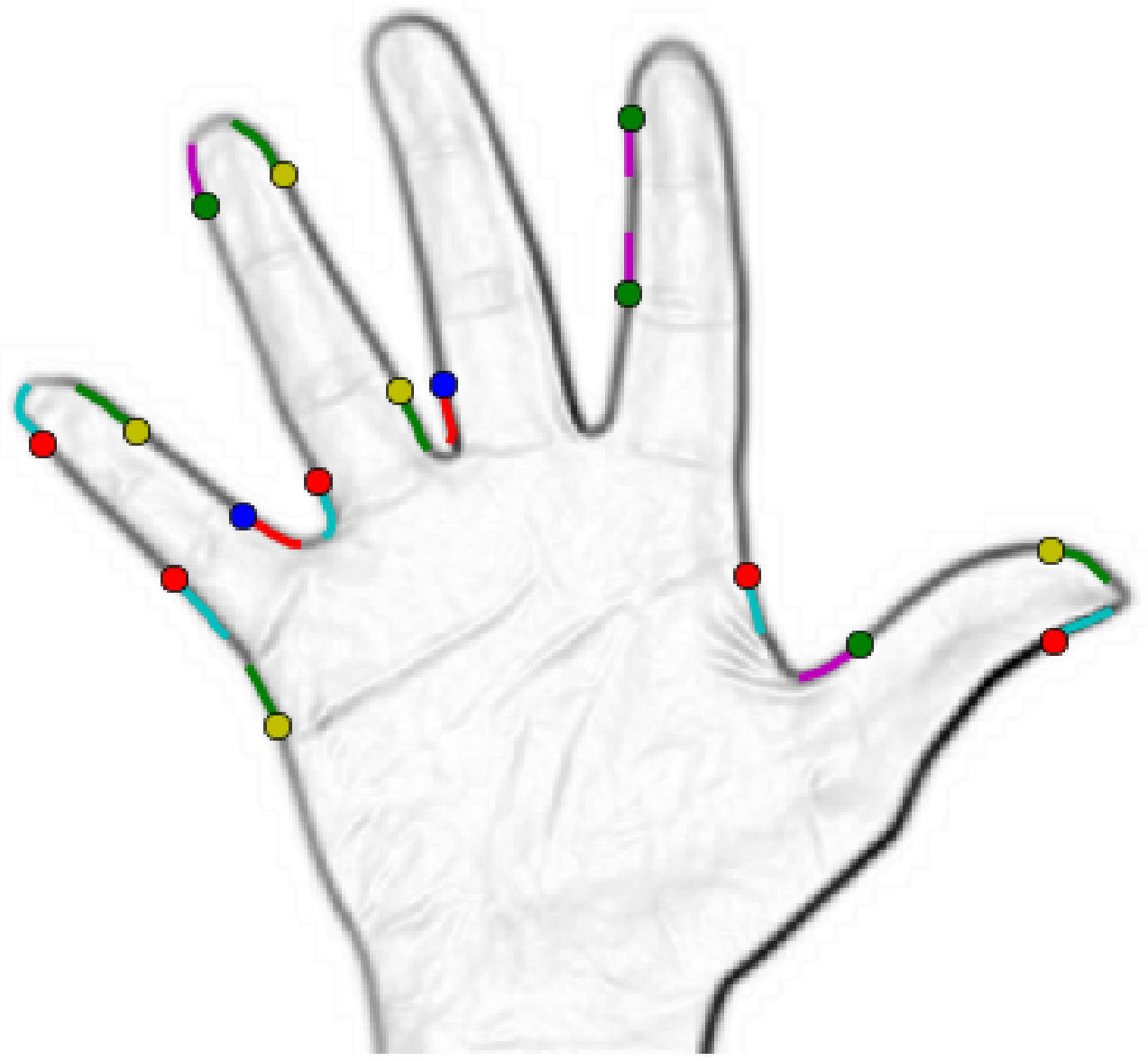,width=0.33\textwidth}
                \epsfig{figure=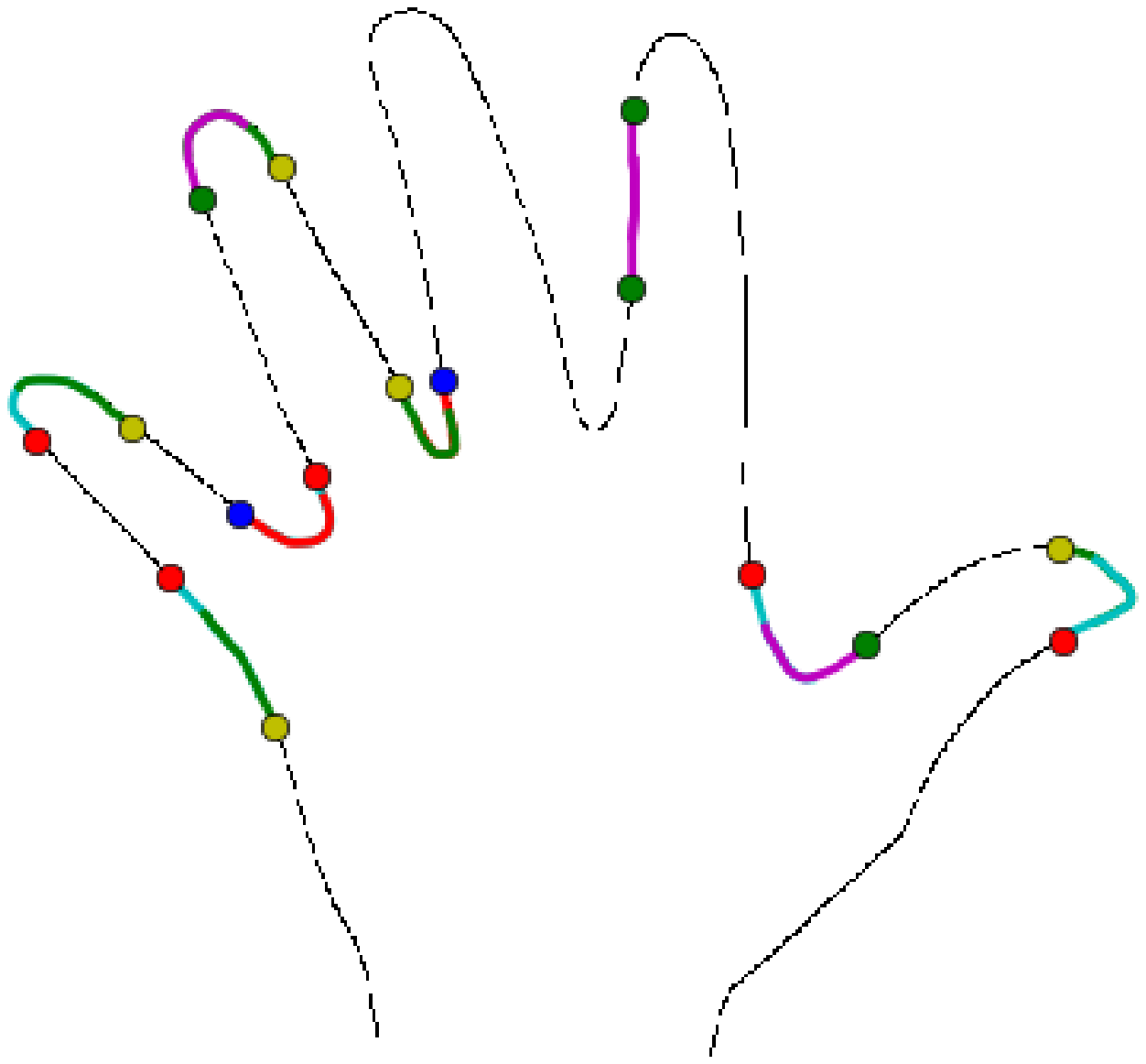,width=0.33\textwidth}}

    \caption{Recovering broken edges with the help of gradient magnitudes computed on the original color image. First rows: edge recovery using only the provided edge image with breaks. Second rows: edge recovery using the gradient magnitudes and edge image with breaks. Initial snakelets are the same in both cases. Using the gradient magnitudes as additional information improves the recovery, the recovered edges are closer to the original edges. Filled circles are starting point of the snakelets.}
    \label{fig:snake-image-recover}
\end{figure}

\subsection{Edge Recovery with the Help of Gradient Magnitudes}

We have discussed edge recovery using only an edge image. The snakelets try to capture the local shape of the existing edge pixels and using only this information grow and recover the missing edge pixels. An intuitive question may follow: can we do better if we also have the original gradient magnitudes, or the original image from which they can be computed? Gradient magnitudes provide additional information, especially across the breaks; this is unavailable in the edge image.

When gradient magnitudes are available, we use them to compute the GVF to better guide the snakelets along the contours as they grow. We use the nonmaximum suppressed (NMS) gradient magnitudes to compute the GVF. However, normalization of GVF causes the noisy gradient magnitudes to amplify and adversely affect the snakelets. As a remedy, we set 0.2-fractile of GVF magnitudes to zero before normalization, to eliminate very low values.

As before, snakelets are initialized at the end points of the breaks. Then, they are deformed and grown under the influence of the GVF. Figure~\ref{fig:snake-image-recover} shows two examples, comparing edge recovery using only a binary edge image and using a binary edge image with its gradient magnitudes. As expected, the final snakelets follows the object contours better with the help of gradient magnitudes (e.g., see finger tips, bunny's ear).

\section{Edge Detection with Snakelets}
\label{sect:ed-snakelets}

In the previous section, we presented an edge recovery algorithm using snakelets; the input was a binary edge image with breaks. In this section, we present a Canny-like edge detection algorithm using snakelets; the input is a color or grayscale image. Namely, we replace the hysteresis thresholding and edge linking step of Canny with snakelets to better utilize the local edge shapes and at the same time recover the broken edges, as described above.

\begin{figure}[h!t]
    \centerline{Image \hspace{20mm} GM \hspace{20mm}  NMS \hspace{20mm} Canny}
    \centerline{\epsfig{figure=figures/bunny.eps,width=0.25\textwidth} 
		\epsfig{figure=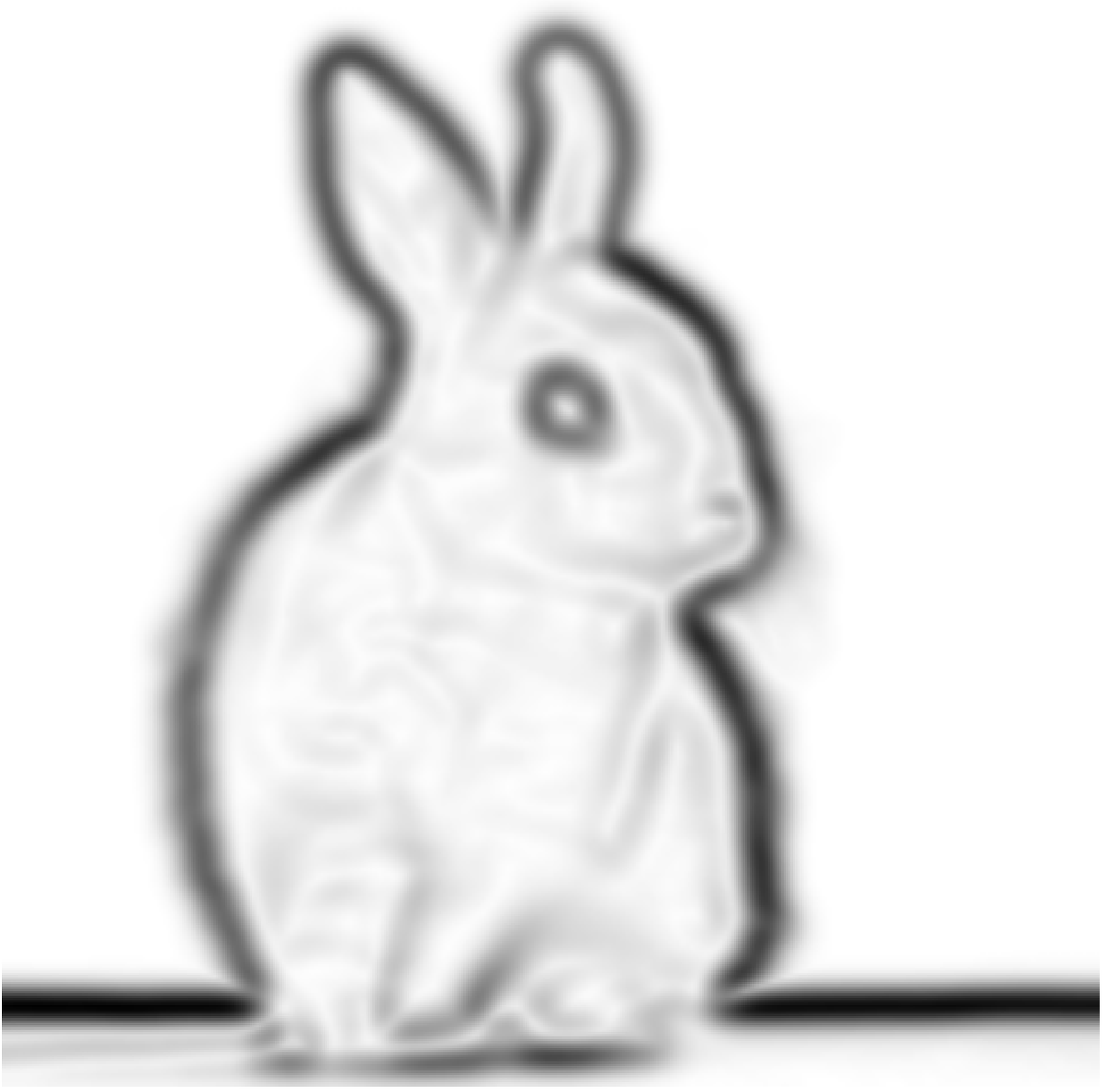,width=0.25\textwidth}
                \epsfig{figure=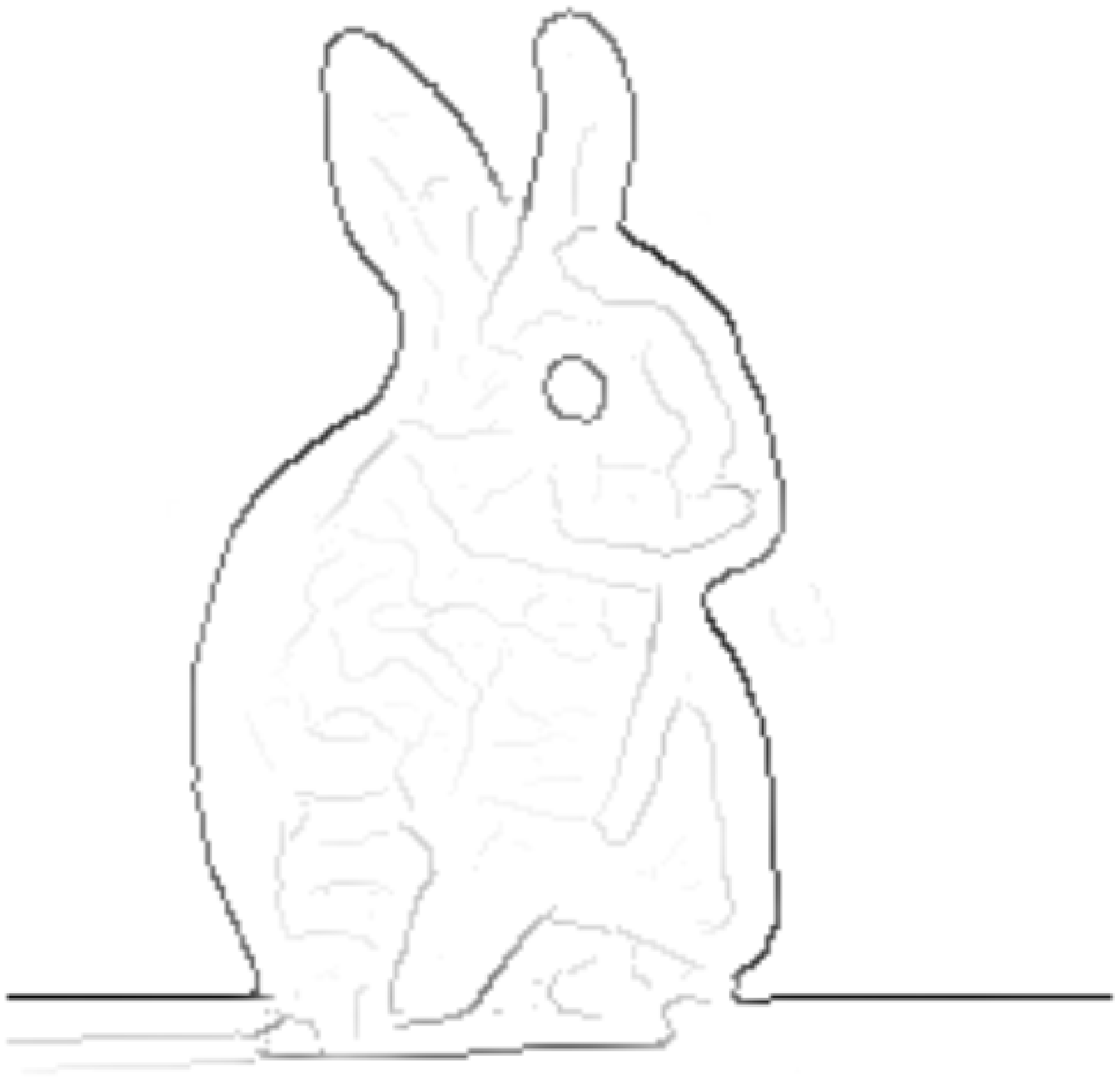,width=0.25\textwidth}
                \epsfig{figure=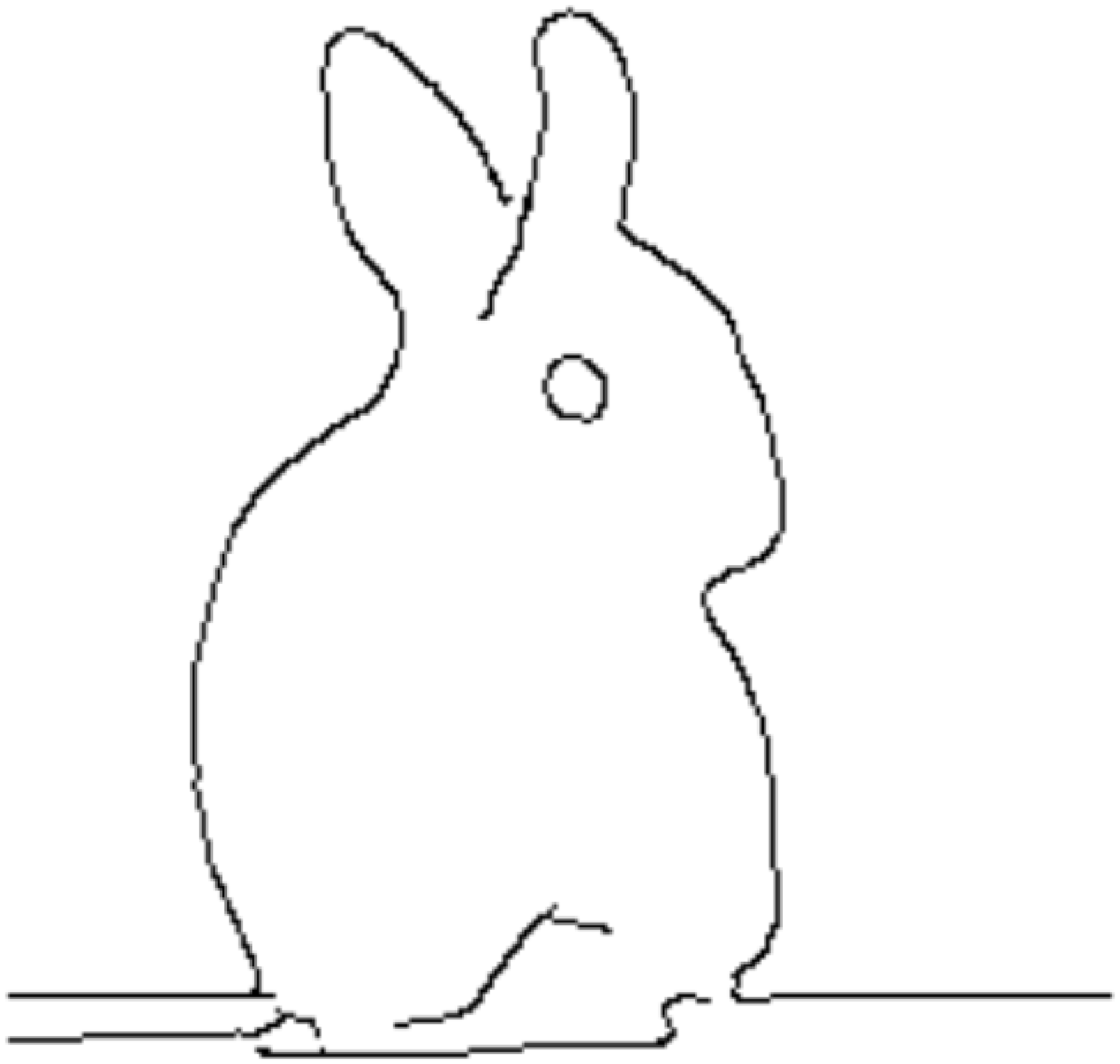,width=0.25\textwidth}}
    \vspace{4mm}
    \centerline{Seed pixels \hspace{10mm} Initial snakelets \hspace{10mm}  Final snakelets \hspace{10mm} Final snakelets}
    \centerline{\epsfig{figure=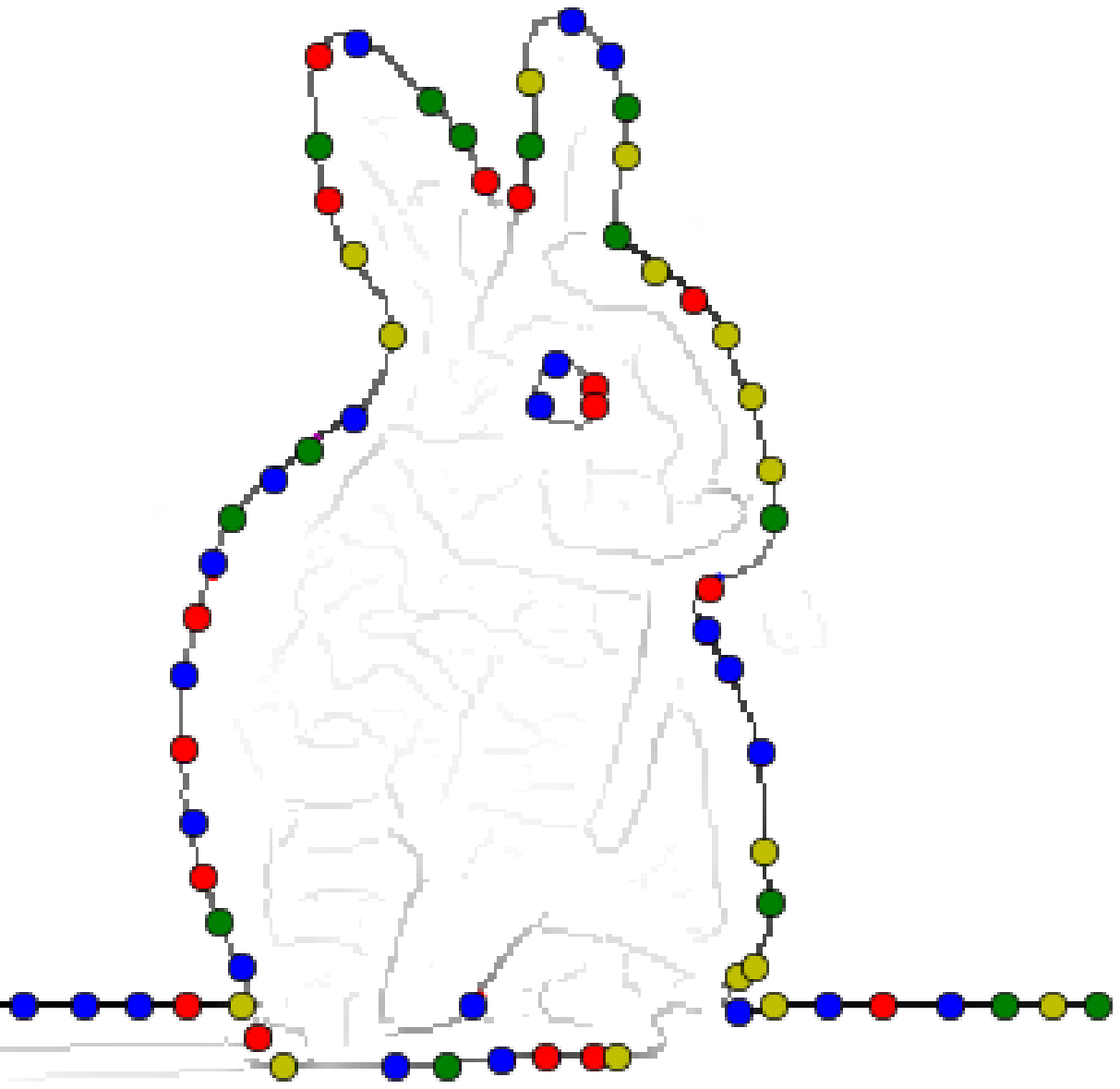,width=0.25\textwidth} 
		\epsfig{figure=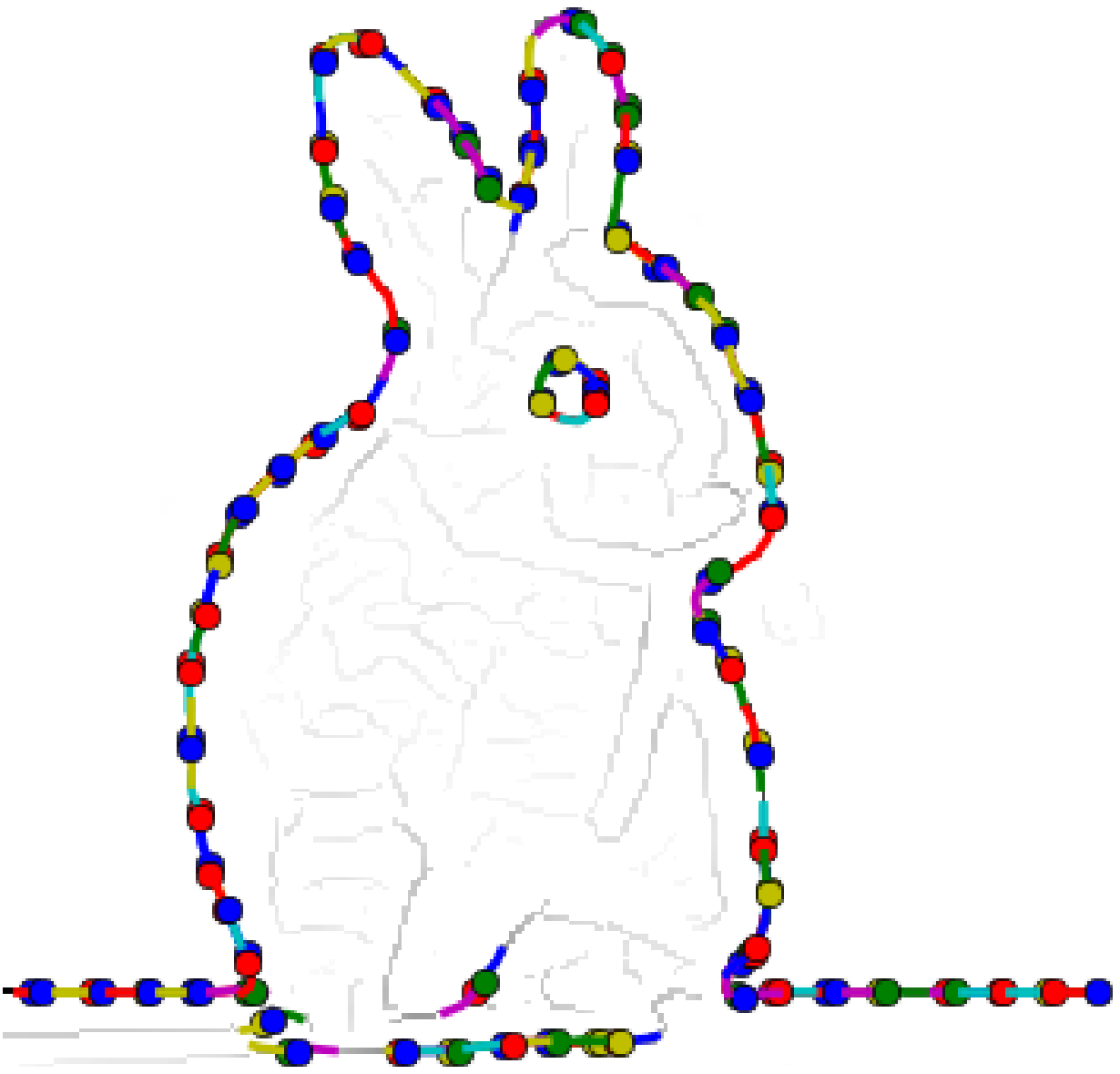,width=0.25\textwidth}
                \epsfig{figure=figures/bunny-trace-final-snakes.eps,width=0.25\textwidth}
                \epsfig{figure=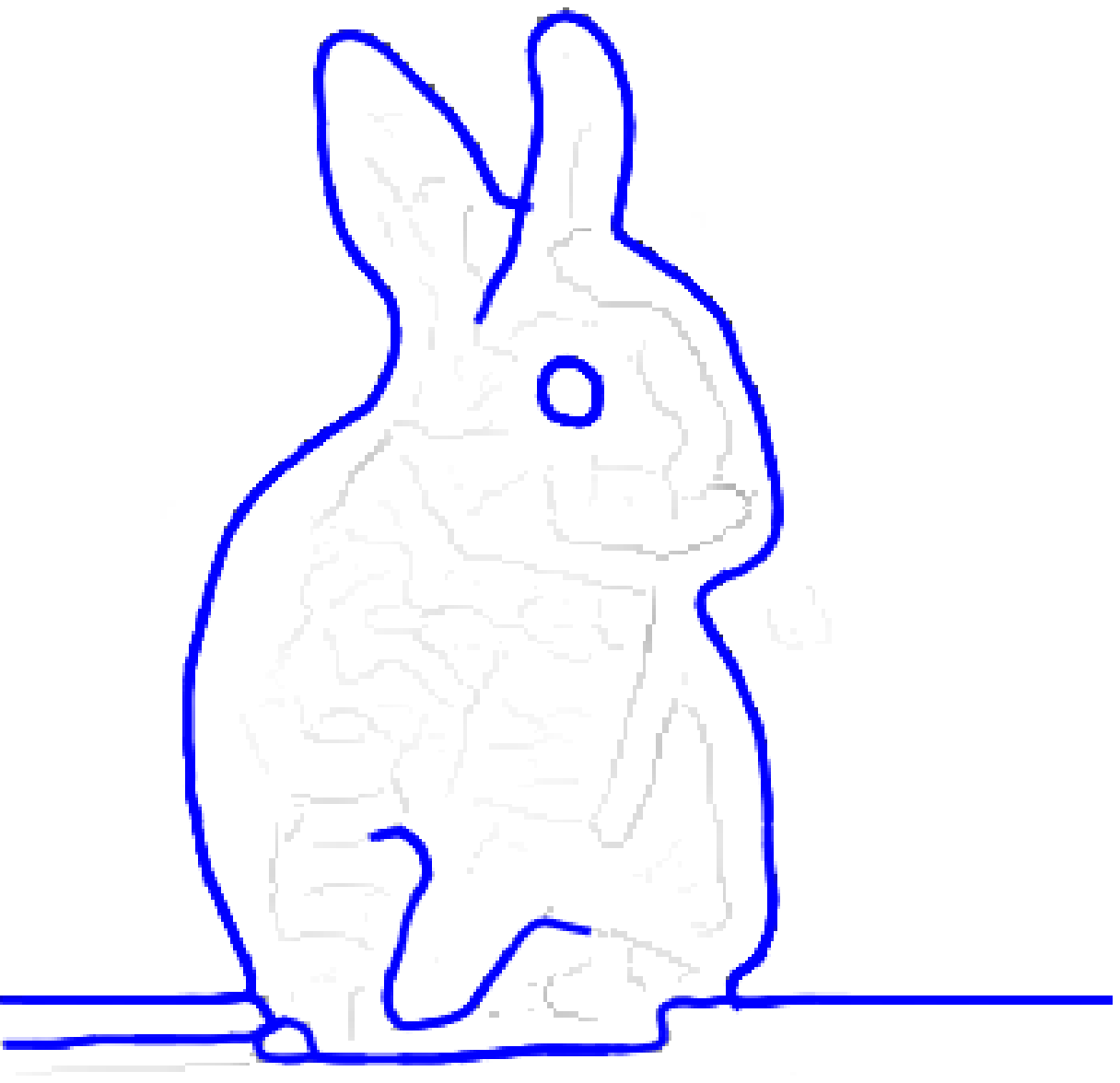,width=0.25\textwidth}}
          
    \caption{Edge detection with snakelets. GM: gradient magnitude image, NMS: non-maximum suppressed gradient magnitude image.}
    \label{fig:snake-ed}
\end{figure}

Figure~\ref{fig:snake-ed} summarizes the major steps of edge detection with snakelets. First, gradient magnitudes are computed on the grayscale or color~\cite{weijer-tip-2006} image. The gradient magnitudes are nonmaxima suppressed. Similar to Canny, we use a high threshold (TH) and a low threshold (TL). Pixels having gradient magnitudes higher than TH are selected as seeds (second row, first image), from which snakelets are initiated in decreasing order of gradient magnitude (second row, second image). Two unidirectional snakelets are initiated from a seed and they grow in opposite directions.

The initial snakelets grow to a specific initial length, e.g., $10-15$. The GVF is computer over the nonmaximum suppressed gradient magnitude image with $3-5$ initial iterations, as before. If a seed pixel is already covered by a previous snakelet, it is discarded. This way, we obtain a set of short snakelets over the strong gradients. Then, the snakelets grow and split recursively until they reach an existing snakelet or they cannot grow anymore (gradient lower than TL). When a snakelet reaches maximum growing length, e.g., $40$, it stops growing; a new short snakelet is initiated at the end of it to continue growing. The growing force is proportional to the gradient magnitude at the end of the snakelet. This recursive growing of snakelets is analogous to the hysteresis thresholding and edge linking of Canny edge detection. However, a snakelet grows in one direction, while edge pixels are linked in all directions in Canny.

\begin{figure}[h!t]
    \centerline{Image \hspace{30mm} Canny \hspace{30mm}  Snakelets}
    \centerline{\epsfig{figure=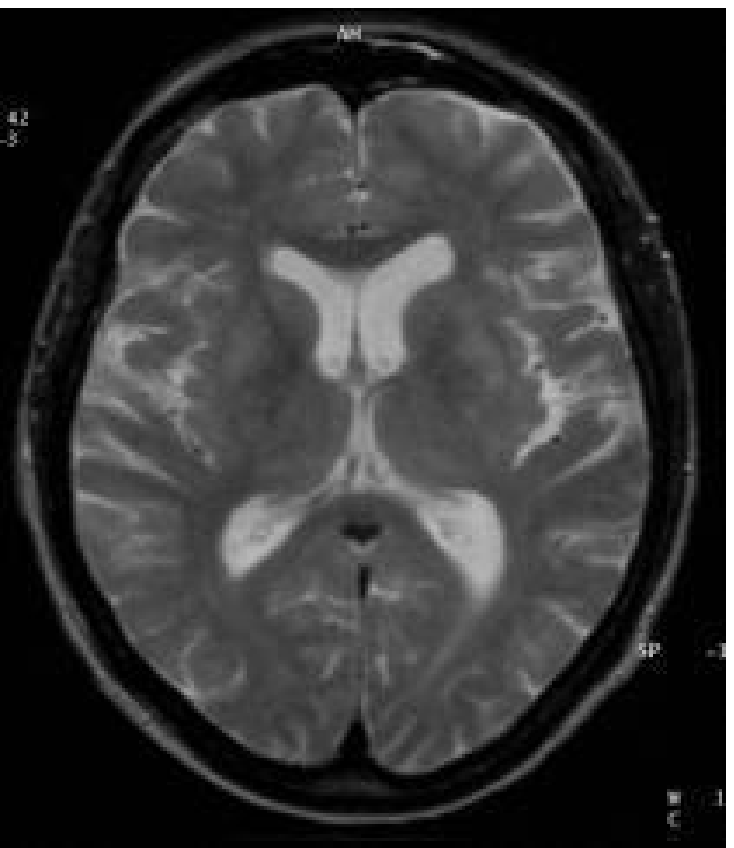,width=0.33\textwidth} 
		\epsfig{figure=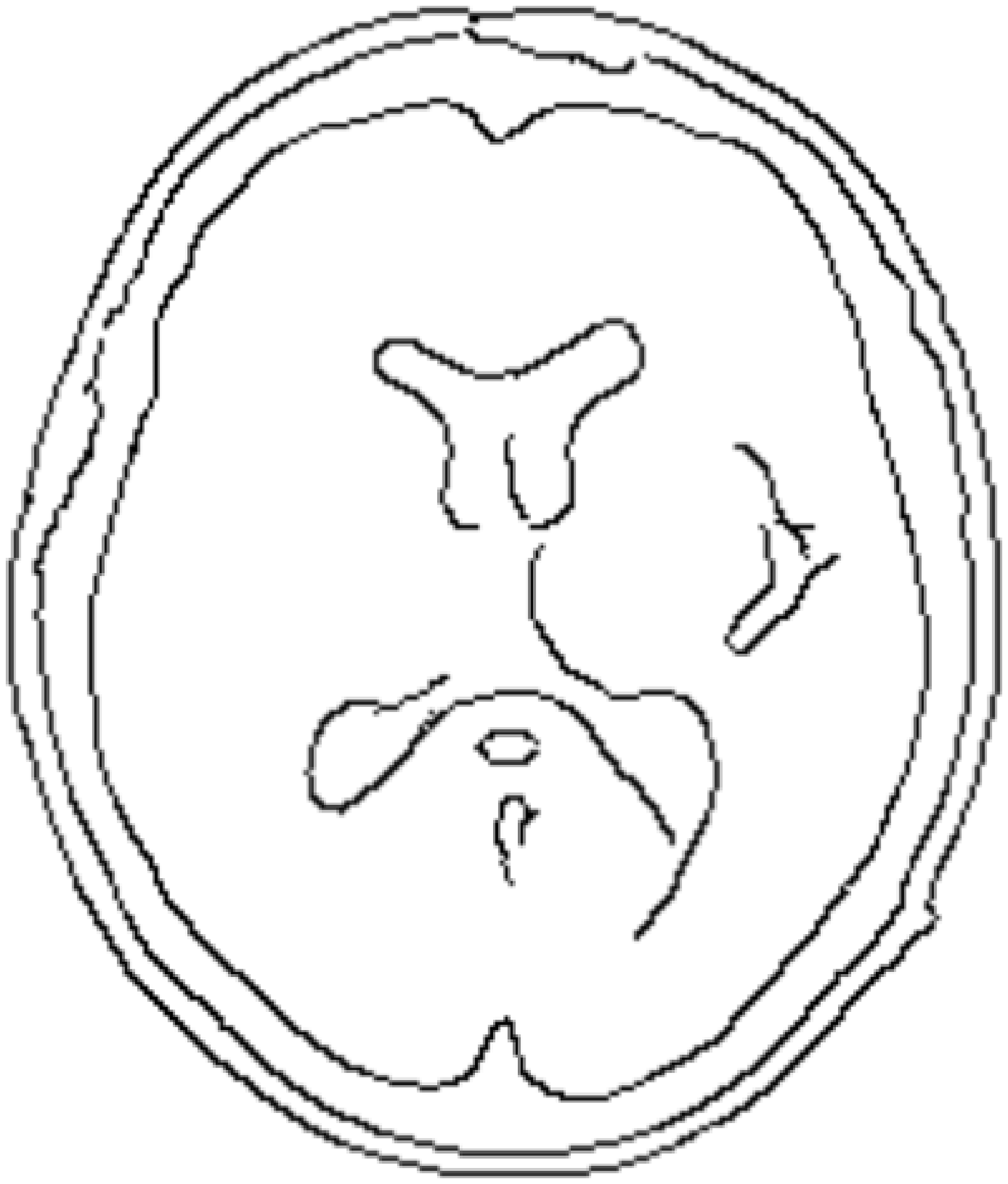,width=0.33\textwidth}
                \epsfig{figure=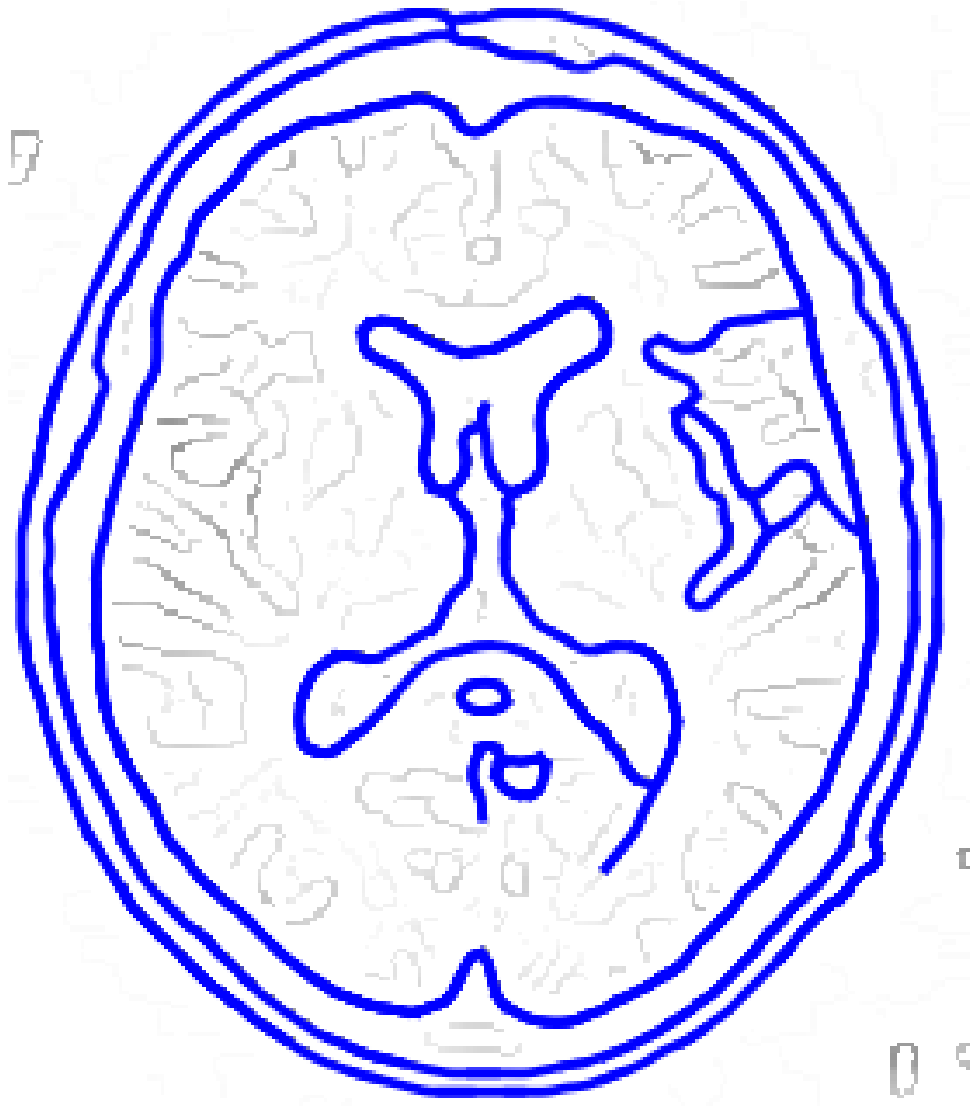,width=0.33\textwidth}}
    \vspace{4mm}
    \centerline{\epsfig{figure=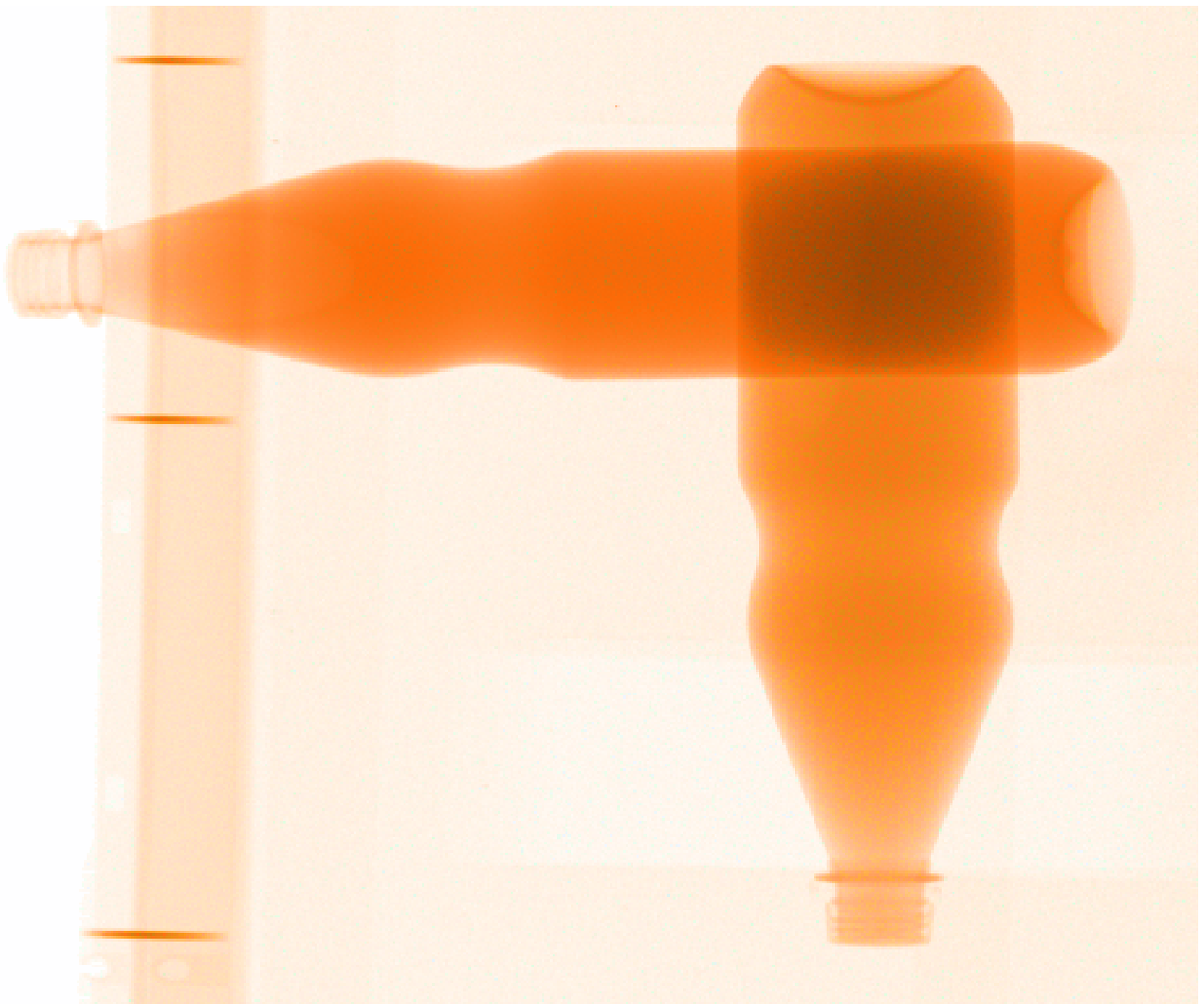,width=0.33\textwidth} 
		\epsfig{figure=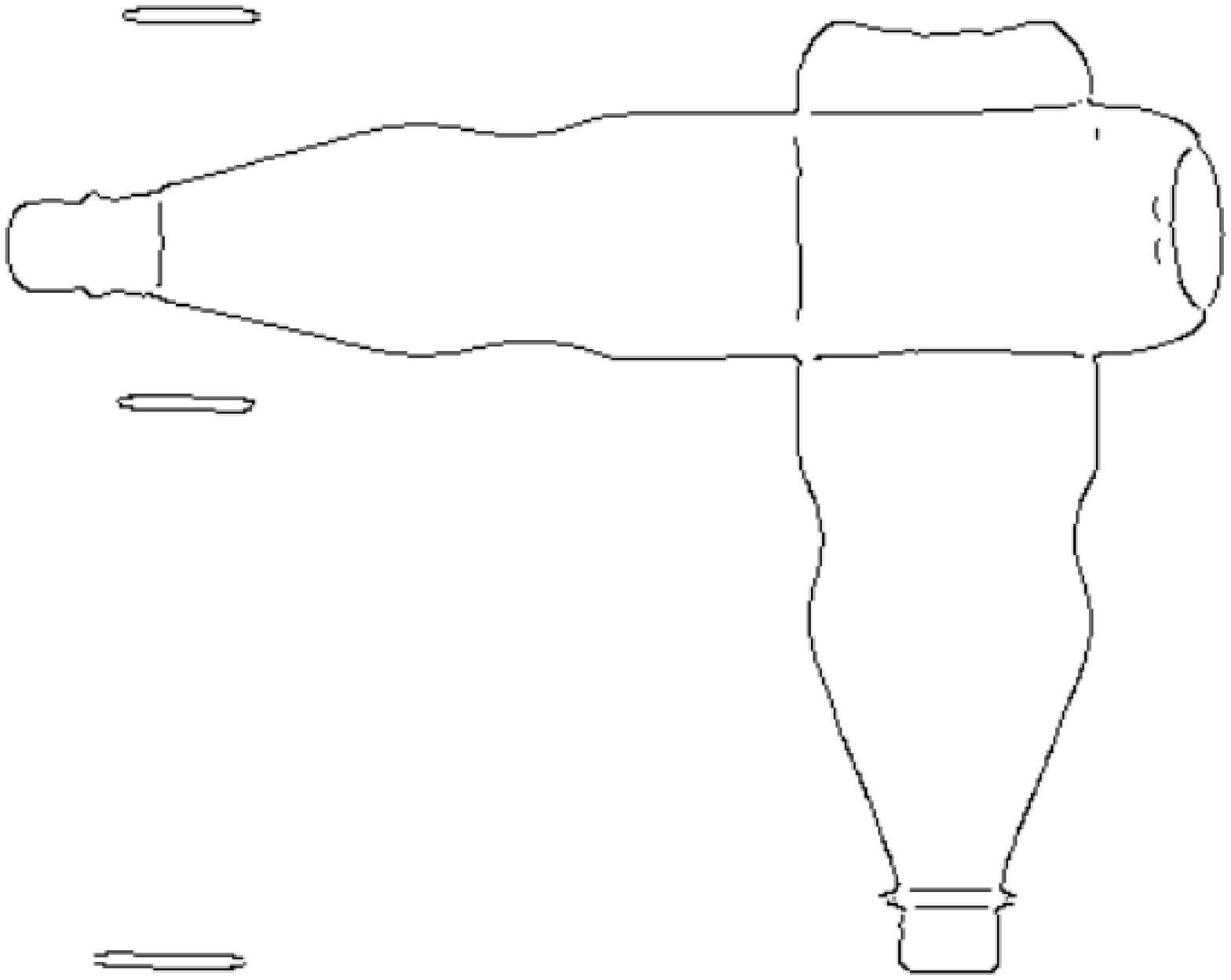,width=0.33\textwidth}
                \epsfig{figure=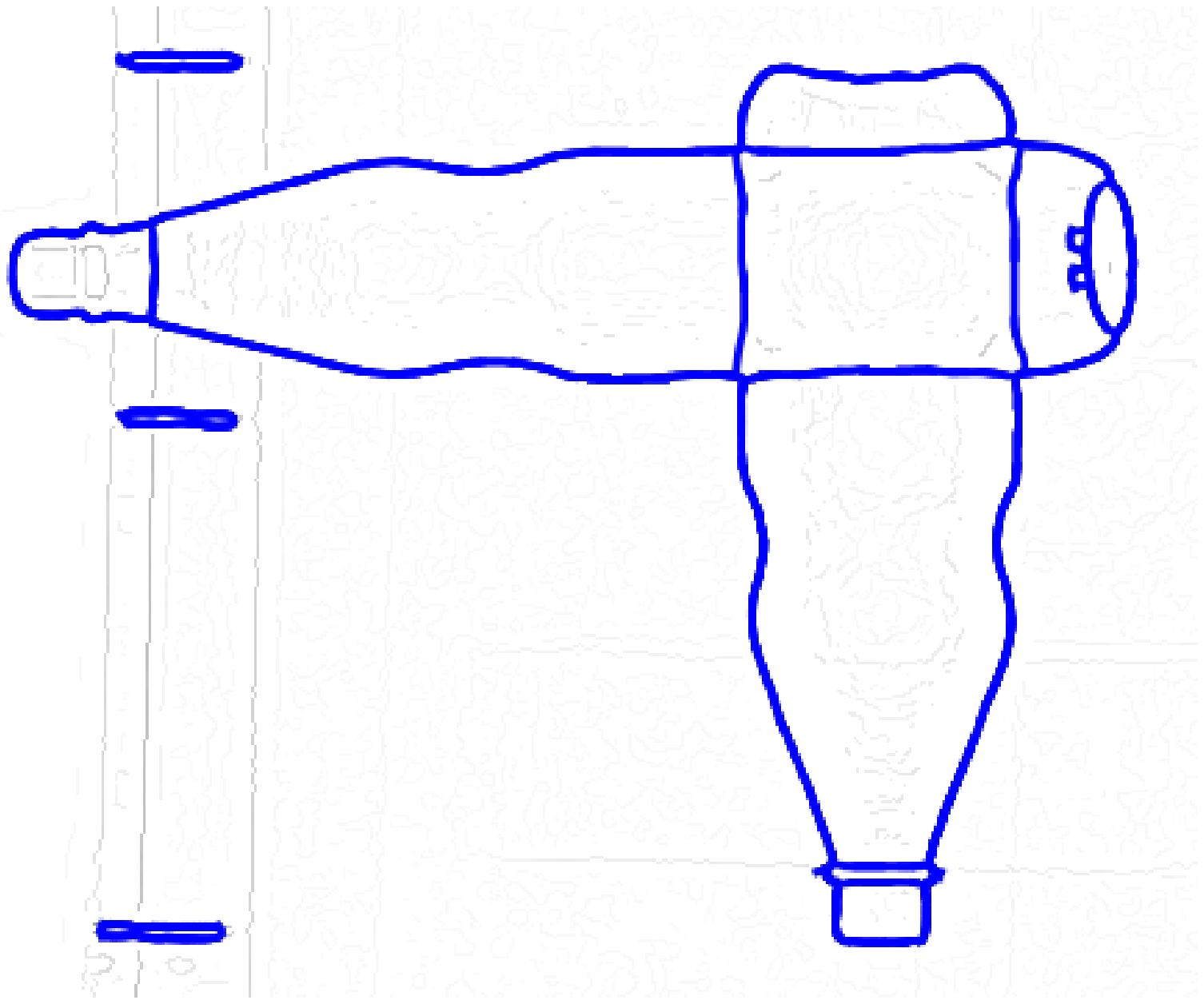,width=0.33\textwidth}}
    \vspace{4mm}
    \centerline{\epsfig{figure=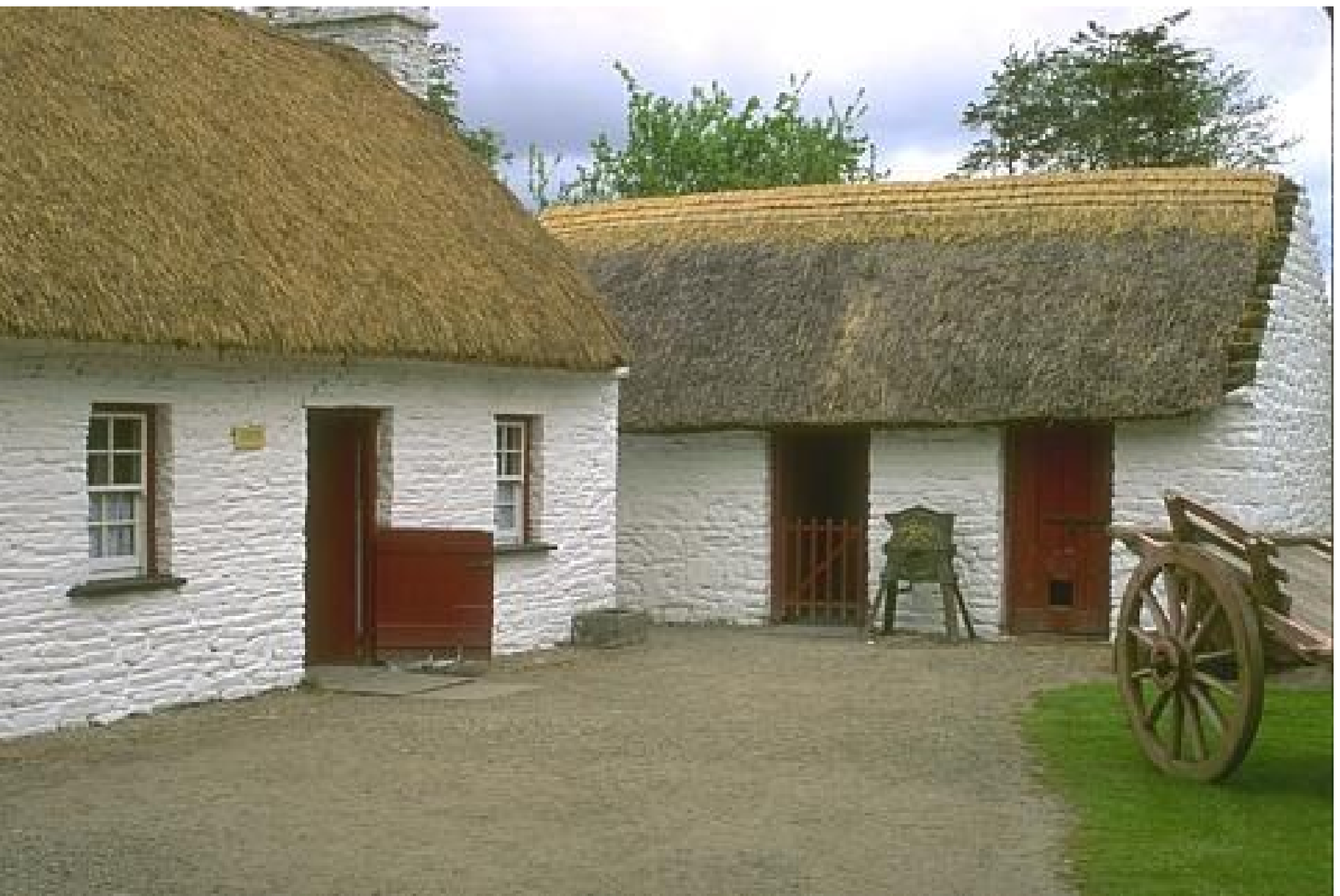,width=0.33\textwidth} 
		\epsfig{figure=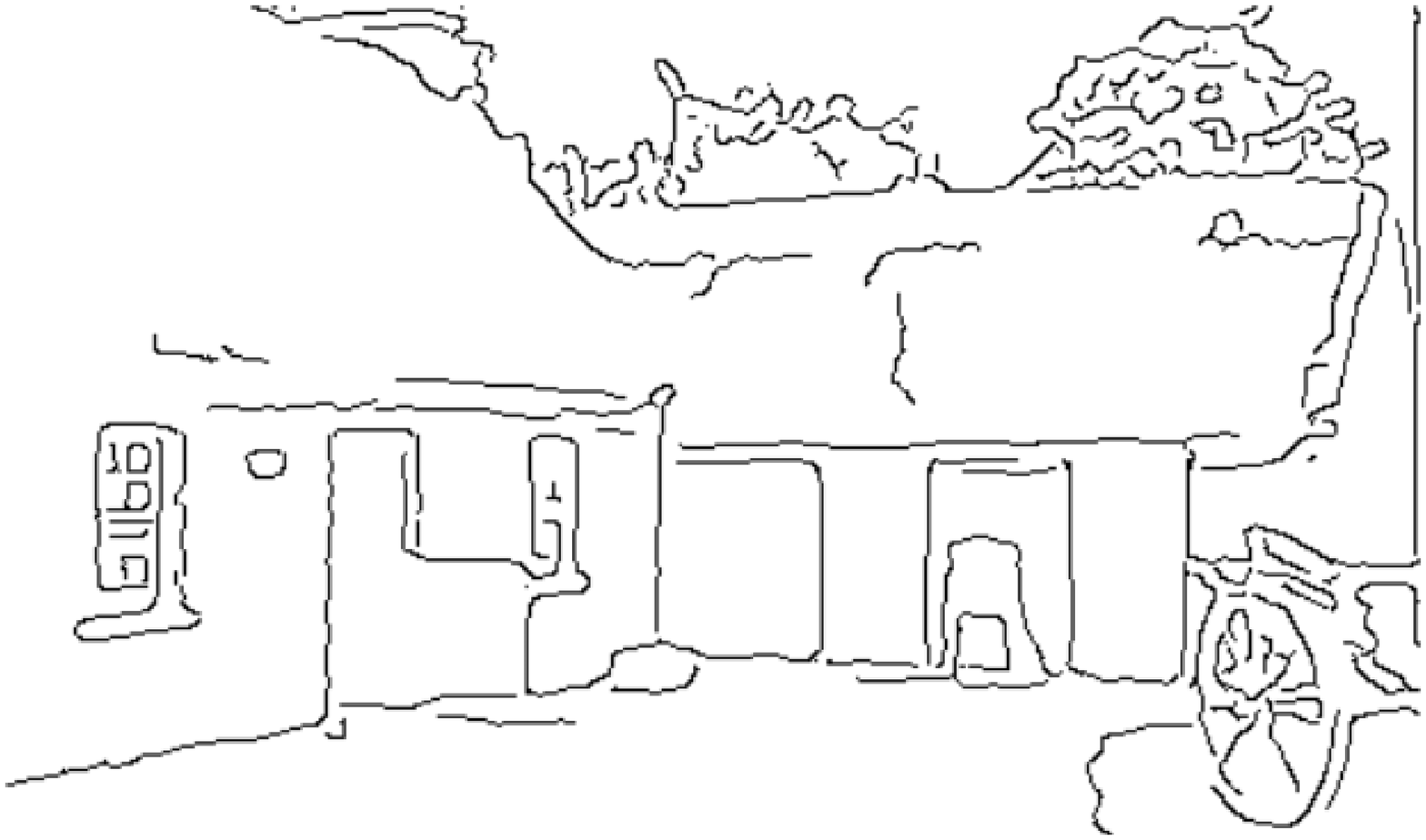,width=0.33\textwidth}
                \epsfig{figure=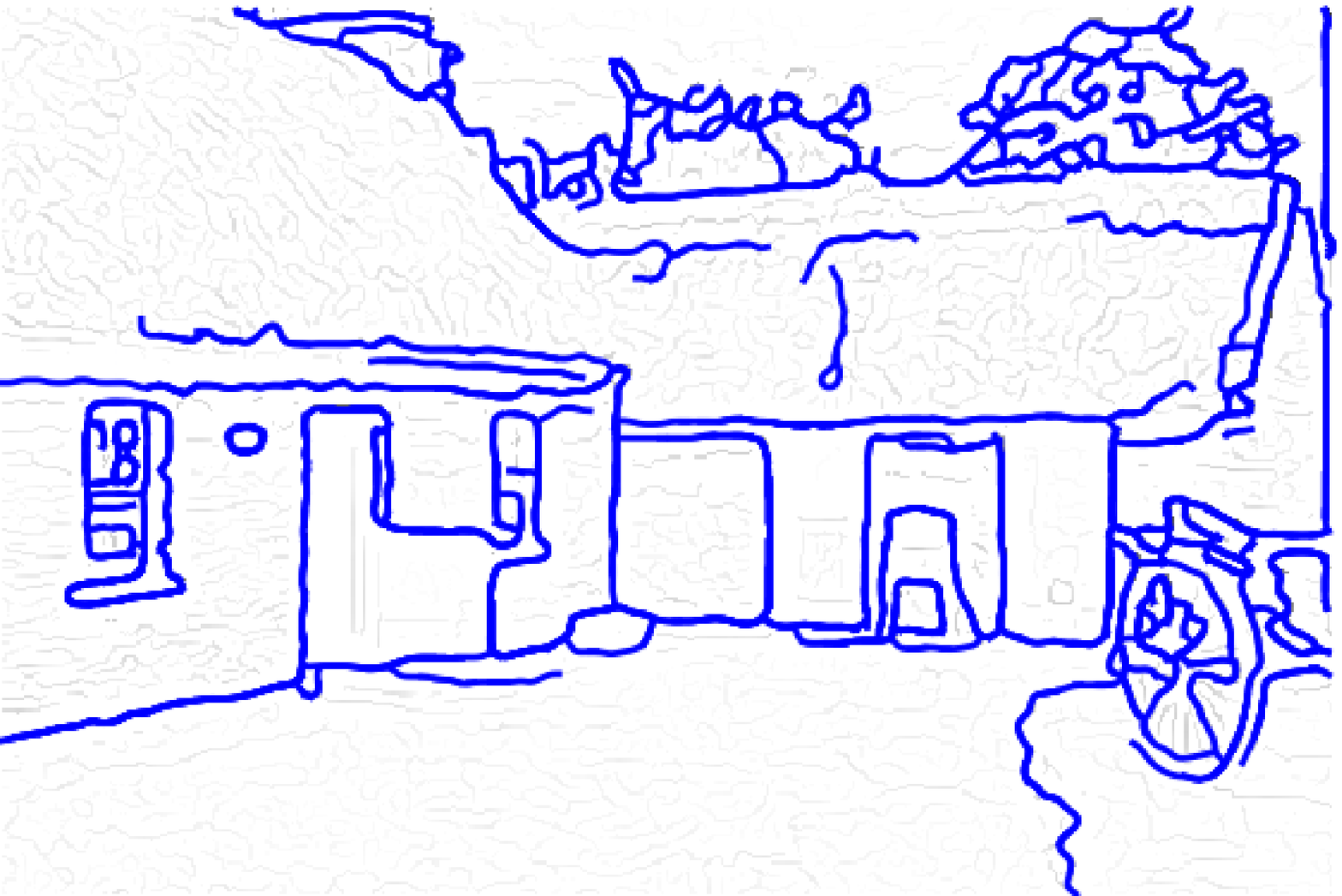,width=0.33\textwidth}}
    \vspace{4mm}
    \centerline{\epsfig{figure=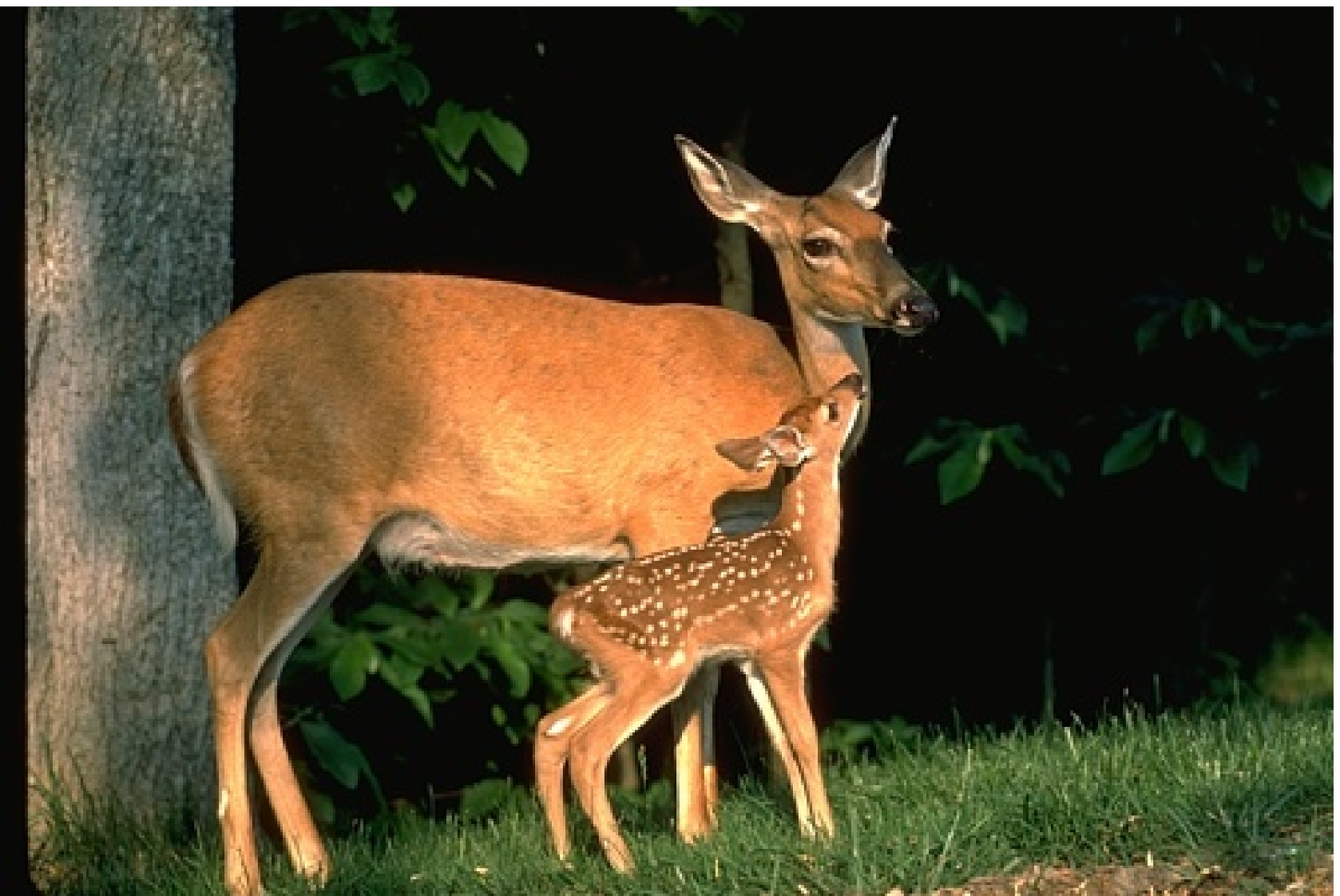,width=0.33\textwidth} 
		\epsfig{figure=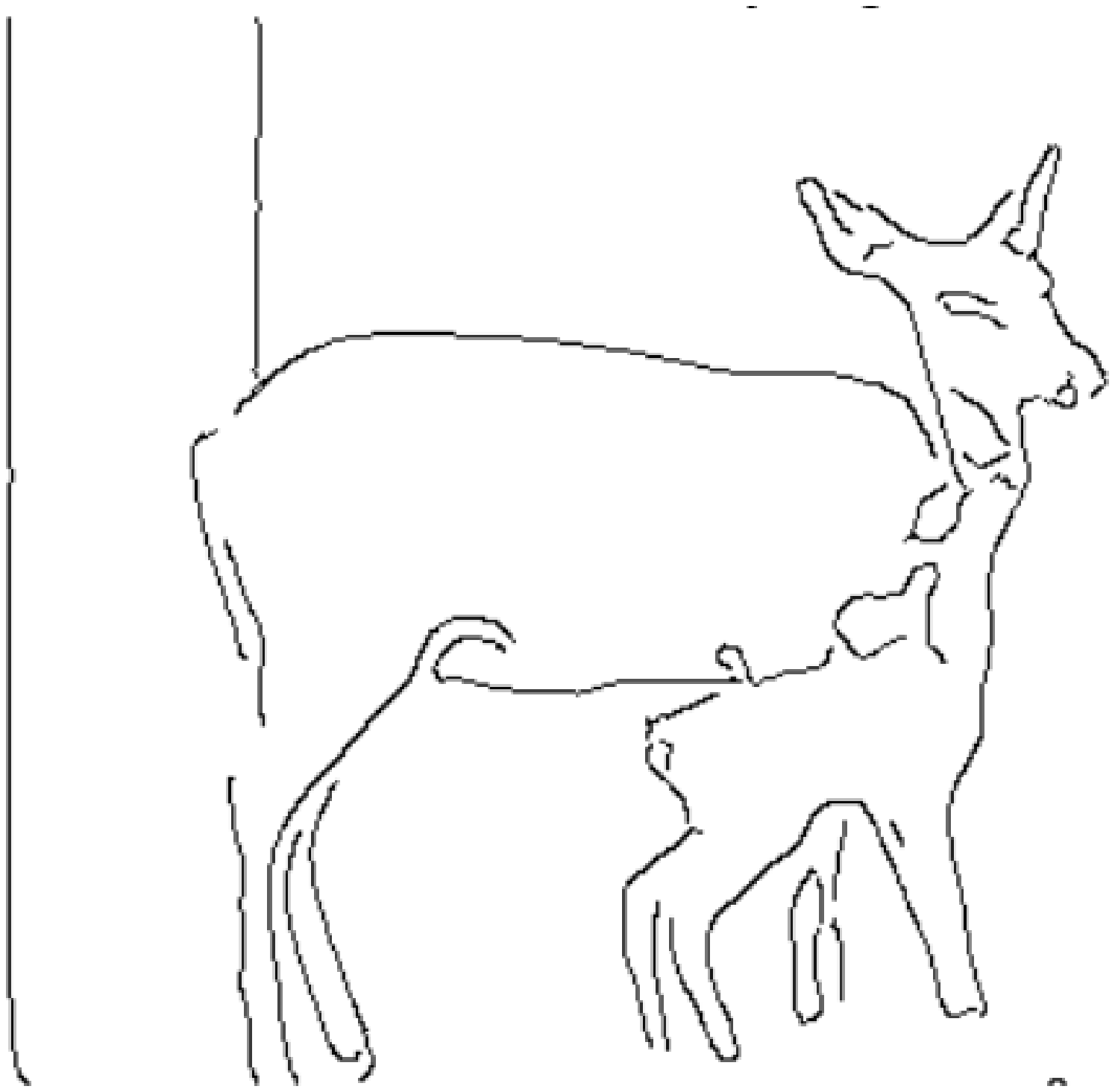,width=0.33\textwidth}
                \epsfig{figure=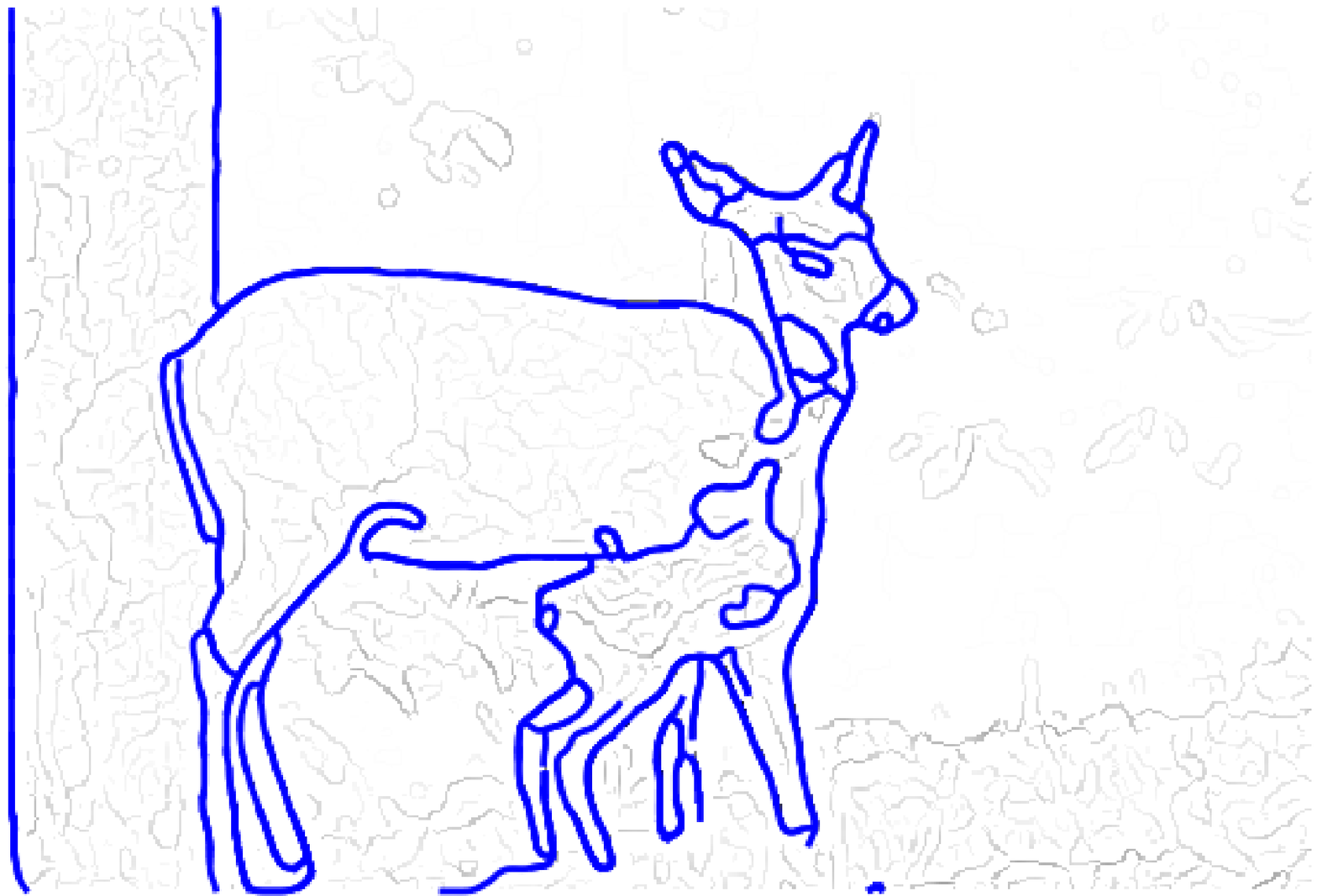,width=0.33\textwidth}}
    \vspace{4mm}
        
    \caption{Canny edge detection vs. snakelet edge detection. Left: input image, middle Canny edges, right: snakelets edges (all snakelets plotted in blue, overlaid the NMS image). First row: brain MR image, second row: baggage X-ray image, third and forth rows: RGB images from the Berkeley Segmentation Dataset (BSD)~\cite{bsd300}.}
    \label{fig:samples-sed}
\end{figure}

Finally, the recovery algorithm is run on snakelets, as described in the previous section, to recover breaks, if any. Final snakelets (second row, last two images) are the outputs and represent the edges in the image.
Figure~\ref{fig:samples-sed} shows sample edge detections with Canny and snakelets using the same TH and TL thresholds, on magnetic resonance (MR), X-ray and RGB color images (the thresholds are the same in the same class of images). The gradients in RGB color images are computed by the algorithm in~\cite{weijer-tip-2006}. Most of the breaks in Canny edges were recovered by snakelets, within limits.

The examples also demonstrate the limitations of snakelets: (1) Snakelets are more suitable for less textured images or for obtaining a coarse outline of objects in images. Snakelets jump across edges in fine textured regions producing spurious edges, as in the trees in the third row. For fine textured images, the initial GVF iterations should be set low to reduce jumping snakelets. For obtaining a coarse outline of objects, the gradient thresholds and GVF iterations should be set high. (2) Snakelets are able to recover breaks if end points are aligned or converging or there is some gradient magnitude to guide them; they are unable to recover breaks if gradient magnitude is near zero across the break and end points are not converging.

The running time of edge recovery or edge detection with snakelets depends on the image size, edge content and actually the number of snakelets. Recovering binary edges is usually faster than direct edge detection with snakelets, since the number of snakelets is less. An unoptimized, single threaded Python implementation takes $1.8$ seconds for edge recovery on the binary bunny image ($225\times224$ pixels, Figure~\ref{fig:snake-recover}, last row). Color Canny edge detection takes $0.32$ seconds, and snakelets edge detection takes about $1.34$ seconds on the $225\times224$ bunny image (Figure~\ref{fig:snake-ed}), and $1.25$ seconds for color Canny and $3.37$ seconds for snakelets on the $481\times321$ resolution image in Figure~\ref{fig:samples-sed}, last row.
The reason for longer running time on edge recovery on the bunny image is the need for expanding and re-growing some snakelets to recover the difficult breaks. The measurements were taken on an Intel Core i7 laptop with 2.4 GHz processors and 32 GB memory.  The snakelets algorithm can easily be parallelized to achieve much higher speeds.

\section{Conclusions}
We have presented an edge detection and recovery framework based on open active contour models.
The output snakelets can be used as the edge representation of the image or they can be used for contour completion or segmentation~\cite{contour-iccv05,contour-ijcv08,contour-cvpr12}, since the snakelets provide cues on the continuity of the contours.

\clearpage
\section{Acknowledgements}
The major part of this work was done when the authors were with the Department of Computer Science, Technical University of Kaiserslautern, Germany. The second author is currently with German Research Center for Artificial Intelligence (DFKI), and third author is with Google.

\bibliographystyle{spmpsci}
\bibliography{References}

\end{document}